\pdfoutput=1
\documentclass[11pt]{article}
\usepackage[final]{acl}

\usepackage{times}
\usepackage{latexsym}
\usepackage[dvipsnames,usenames]{xcolor}
\usepackage{makecell}
\usepackage{pdflscape}
\usepackage[T1]{fontenc}
\usepackage[utf8]{inputenc}
\usepackage{booktabs}
\usepackage{amsmath, amssymb, mathtools}
\usepackage{pdflscape}

\usepackage{microtype}
\usepackage[normalem]{ulem}
\usepackage{inconsolata}
\usepackage{longtable}
\usepackage{subcaption}
\usepackage{graphicx}
\usepackage{tikz}
\usepackage{tikz-dependency}
\usepackage{multirow}
\newcommand{\Enc}{\mathcal{E}}

\newcommand{\best}[1]{\textbf{#1}}
\title{Dependency Graph Parsing as Sequence Labeling}
\author{Ana Ezquerro, David Vilares, Carlos Gómez-Rodríguez \\
  Universidade da Coruña, CITIC \\
  Departamento de Ciencias de la Computación y Tecnologías de la Información \\
  Campus de Elviña s/n, 15071 \\
  A Coruña, Spain\\
  \texttt{\{ana.ezquerro, david.vilares, carlos.gomez\}@udc.es}} 
\definecolor{pumpkin}{HTML}{FE7F2D}
\definecolor{violetblue}{HTML}{454E9E}
\definecolor{seagreen}{HTML}{018E42}
\definecolor{argentinianblue}{HTML}{77B6EA}
\definecolor{robineggblue}{HTML}{06BCC1}
\definecolor{steelblue}{HTML}{3f88c5}

\definecolor{plane1color}{HTML}{0000FF}
\definecolor{plane2color}{HTML}{E62020}
\definecolor{plane3color}{HTML}{018E42}

\newcommand{\planei}[1]{\textcolor{plane1color}{#1}}
\newcommand{\planeii}[1]{\textcolor{plane2color}{#1}}
\newcommand{\planeiii}[1]{\textcolor{plane3color}{#1}}

\begin{document}
\maketitle
\begin{abstract}
Various linearizations have been proposed to cast syntactic dependency parsing as sequence labeling. However, these approaches do not support more complex graph-based representations, such as semantic dependencies or enhanced universal dependencies, as they cannot handle reentrancy or cycles.
By extending them, we define a range of unbounded and bounded linearizations that can be used to cast graph parsing as a tagging task, enlarging the toolbox of problems that can be solved under this paradigm. Experimental results on semantic dependency and enhanced UD parsing show that with a good choice of encoding, sequence-labeling dependency graph parsers combine high efficiency with accuracies close to the state of the art, in spite of their simplicity.
\end{abstract}

\section{Introduction}

In recent years, a new family of approaches has emerged for dependency parsing that treats the problem as a sequence labeling task~\citep{strzyz-etal-2019-viable}. This has advantages in terms of simplicity, flexibility and speed, as parsing can be performed with generic sequence labeling software and easily combined with other tasks that operate within the same framework. For this, one needs an encoding that can represent parse trees as a sequence composed of one discrete label per word, so that a sequence labeling component can be trained and output trees can then be decoded from the sequences.

In the last few years, a wide range of such encodings with different strengths and weaknesses have been proposed for dependency parsing~\citep{strzyz-etal-2019-viable,lacroix-2019-dependency,strzyz-etal-2020-bracketing,gomez-rodriguez-etal-2023-4,amini-etal-2023-hexatagging}. However, these encodings are designed for dependency trees, where each word is restricted to have exactly one parent and cycles are forbidden. The more complex family of structured prediction problems where the output is a graph of dependencies between words, including tasks like semantic dependency parsing~\citep{oepen-etal-2015-semeval}, enhanced Universal Dependencies parsing~\citep{bouma-etal-2021-raw} or even structured sentiment analysis~\citep{barnes-etal-2021-structured}; has not been integrated into the sequence labeling framework so far due to a lack of encodings that can support reentrancy, disconnected nodes and cycles.\footnote{Note that there exist linearizations to implement semantic dependency parsing with sequence-to-sequence (seq2seq) approaches~\citep{lin-etal-2022-dependency}. However, seq2seq models are not to be confused with sequence labeling models. In seq2seq models, output length is arbitrary. For sequence labeling, the output needs to have exactly the same length as the input, i.e., graphs need to be encoded as one label per word. Existing seq2seq linearizations do not meet this condition, so they cannot be used for sequence labeling.}

We bridge this gap by defining sequence labeling encodings for these problems that are framed as predicting directed graphs between words - which, following some previous literature~\citep{agic-etal-2015-semantic,barnes-etal-2021-structured} - we group under the umbrella term of \emph{dependency graph parsing}. By taking dependency tree encodings as a starting point and extending them to support graphs, we define a wide range of both unbounded and bounded encodings to cast dependency graph parsing problems as sequence labeling. To test the performance of the encodings, we experiment on two dependency graph parsing tasks, representative of different kinds of linguistically-relevant structures: semantic dependency parsing (where the output are DAGs, directed acyclic graphs) and enhanced UD parsing (where graphs have cycles). The source code is available at \href{https://github.com/anaezquerro/separ}{\texttt{github.com/anaezquerro/separ}}.

\section{Background}\label{sec:background}

Sequence-labeling approaches that partially perform syntactic parsing have long been known, in the form of supertagging~\citep{joshi-srinivas-1994-disambiguation}. Still, the goal of supertagging is to cut the search space of the parsing process, not to fully replace it: the generated labels (supertags) are not enough to encode a whole parse tree, and a parsing algorithm is still needed to fill the missing information.

The first attempt at addressing the full task of natural language parsing as sequence labeling was by~\citet{spoustova2010dependency}, who introduced a linearization for dependency parsing where the label of each word encoded the PoS tag of its head and its relative position among words with the same PoS tag. However,  machine learning algorithms of the time struggled to predict such labels, leaving the practical results far behind the state of the art.

It was with the development of deep learning and its use in NLP architectures that parsing as sequence labeling became practically viable. This was shown by~\citet{gomez-rodriguez-vilares-2018-constituent} for constituent parsing and by~\citet{strzyz-etal-2019-viable} for dependency parsing. For the purposes of this paper, we will leave work on constituent parsing linearizations \cite{kitaev-klein-2020-tetra,amini-cotterell-2022-parsing} aside and outline the encodings that cast dependency parsing as sequence labeling, since they have a more direct relation to our target problem of dependency graph parsing and will be the inspiration of our proposed encodings.

\paragraph{Common notation} 
Let $V$ be a vocabulary of possible tokens. We will denote a sentence of length $n$ by $w = [ w_1, w_2, \ldots, w_n] \in V^n$.
Let $S_n$ be the set of possible parses (dependency trees or dependency graphs) for sentences of length $n$. Then, a \emph{sequence labeling encoding} for parsing is an injective function $\Enc: S_n \rightarrow L^n$; where $L$ is a set of labels that is defined depending on the encoding. Thus, a sequence labeling encoding is used to represent a parse for a sentence of length $n$, $w = [ w_1, w_2, \ldots, w_n]$, as a sequence of $n$ labels (one per word), $l = [ l_1, l_2, \ldots, l_n] \in L^n$. 

Since $\Enc$ is injective, one can recover a parse in $S_n$ from its associated label sequence in $L^n$ via the inverse function, ${\Enc}^{-1}: \{ {\Enc}(s) \mid s \in S_n \} \rightarrow S_n$. This enables parsing as sequence labeling: if we train a tagger to predict a function $f_{\Theta}: V^n \rightarrow L^n$ (where $V$ is the vocabulary of possible input tokens, and $\Theta$ are the model parameters) that associates each word $w$ with the encoding of its parse, we can obtain the parse for a sentence $w$ as ${\Enc}^{-1}(f_{\Theta}(w))$.   

A theoretical limitation is that no known encodings are bijective, so ${\Enc}^{-1}$ is not defined on all possible sequences of labels ($L^n$), but just on those that correspond to encodings of valid trees. We call the set of such sequences, ${\Gamma}_n = \{ {\Enc}(s) \mid s \in S_n \} \subseteq L^n$, the set of \emph{well-formed} label sequences for length $n$. Since $f_{\Theta}$ is an approximation by a sequence labeling classifier, it is possible that it outputs ill-formed label sequences. However, this is workable in practice, since there are simple heuristics to fix ill-formed sequences converting them to well-formed ones (i.e., mapping from $L^n$ to ${\Gamma}_n$).

We now define concepts related to $k$-planarity~\citep{yli2003multiplanarity} to later define the coverage of various encodings. Two arcs in a dependency tree or graph are said to \emph{cross} if their arrows cross when drawn above the words. Thus, two arcs $(w_i,w_j)$ and $(w_k,w_l)$ such that $\min(i,j) < \min(k,l)$ cross iff $\min(k,l) < \max(i,j) < \max(k,l)$. A tree or graph is \emph{noncrossing}, or \emph{1-planar}, if it contains no crossing arcs. We also introduce the term \emph{relaxed 1-planar} for a tree or graph with no pair of crossing arcs pointing in the same direction (i.e., only opposite crossing arcs are allowed). Finally, a tree or graph is \emph{(relaxed) k-planar} for some $k \ge 0$ if it can be written as the union of $k$ (relaxed) 1-planar subgraphs (called planes).

\paragraph{Dependency parsing encodings}
A \emph{sequence labeling encoding for dependency parsing} is a sequence labeling encoding where the set of parses of interest, $S_n$, is $T_n$, the set of dependency trees for sentences of length $n$.

In all dependency parsing encodings defined so far in the literature, each label $l_i$ assigned to a word $w_i$ is of the form $(d_i,x_i)$, where $d_i$ represents the label of the dependency going to $w_i$, and it is $x_i$ that varies between encodings and encodes the unlabeled dependency tree. Thus, we will focus on $x_i$ and ignore dependency labels from now on.

\paragraph{Positional encodings} We call \emph{positional encodings} those where $x_i$ encodes the position of the head of $w_i$. Let $w_h$ be the head of $w_i$. The simplest such encoding is the \emph{naive positional encoding} where $x_i = h$, i.e., it encodes directly the position of the head of $w_i$ (as in the CoNLL format). However, this encoding has been shown to not work well in practice~\citep{strzyz-etal-2019-viable}. Instead, the \emph{relative positional encoding}~\citep{li-etal-2018-seq2seq,strzyz-etal-2019-viable} represents a relative offset, $x_i = h-i$. While it did not obtain good results under simpler implementations, it has been shown to be viable when coupled with more powerful language models~\citep{vacareanu-etal-2020-parsing}.

To reduce sparsity, one can use PoS tags to locate head words. In the \emph{relative PoS-based encoding}~\citep{spoustova2010dependency,strzyz-etal-2019-viable}, $x_i$ is a pair $(p_i,o_i)$ such that if $o_i > 0$ then $w_h$ is the $o_i$th among the words with PoS tag $p_i$ that are located to the right of $w_i$, and if $o_i < 0$, then it is the $-o_i$th among words with PoS tag $p_i$ to the left of $h_i$. This encoding has been shown to be very effective in high-resource setups where high-accuracy PoS tags are avaible~\citep{strzyz-etal-2019-viable}, but tends to suffer when this is not the case~\citep{munoz-ortiz-etal-2021-linearizations}. It is also possible to restrict offsets using properties other than PoS tags, as in the \emph{relative head-based} encoding of~\citet{lacroix-2019-dependency}, based on tagging words as leaf or non-leaf nodes and then encoding $x_i = o_i$ and finding the $o_i$th non-leaf to the right or $-o_i$th to the left.

\paragraph{Unbounded bracketing encodings} Based on the axiomatization by~\citet{yli-jyra-gomez-rodriguez-2017-generic}, these representations encode each dependency arc by adding one symbol to the label of each of its endpoints. In the simplest version, the \emph{basic bracketing encoding} adapted to sequence labeling by~\citet{strzyz-etal-2019-viable}, a right arc from $w_i$ to $w_j$ is represented by including a \planei{\texttt{/}} symbol at the label $x_i$ and a \planei{\texttt{>}} symbol at $x_j$, whereas a left arc from $x_j$ to $x_i$ is encoded by a \planei{\texttt{<}} symbol at $x_i$ and a \planei{\texttt{\textbackslash}} symbol at $x_j$.\footnote{In~\citep{strzyz-etal-2019-viable,strzyz-etal-2020-bracketing}, unbounded bracketing encodings are defined differently, with arcs involving $w_i$ and $w_j$ being encoded at labels $x_{i+1}$ and $x_j$. We choose the straightforward $x_i$ and $x_j$, as the reason for the first option (reducing sparsity in projective trees) is not relevant for this work. } The label for a word is a string formed by concatenating all symbols involving that word, so that for example, a label $x_i = $ \planei{\texttt{\textbackslash>//}} means that the word $w_i$ has one outgoing arc to the left, two to the right, and one incoming arc from the left.

Decoding is a linear-time process where the sentence is read from left to right. Two separate stacks are used: one to decode \planei{\texttt{/}} and \planei{\texttt{>}} into right arcs, and the other to decode \planei{\texttt{<}} and \planei{\texttt{\textbackslash}} into left arcs. Symbols are treated as brackets, with left (opening) brackets \planei{\texttt{/}} and \planei{\texttt{<}} being pushed into the stack when read, and popped when a matching right bracket (respectively, \planei{\texttt{>}} or \planei{\texttt{\textbackslash}}) is found, while the corresponding arc is created. Since \planei{\texttt{>}} is always matched to the closest \planei{\texttt{/}} and \planei{\texttt{<}} to the closest \planei{\texttt{\textbackslash}}, the encoding cannot handle trees that have crossing arcs within the same direction, being restricted to relaxed 1-planar trees.

To improve coverage, one can apply the notion of multiplanarity~\citep{yli2003multiplanarity}, dividing the dependency tree into two separate subgraphs (planes) and encoding each separately. This yields the \emph{2-planar bracketing encoding}~\citep{strzyz-etal-2020-bracketing}, an encoding with two sets of brackets: the original \planei{\texttt{/}}, \planei{\texttt{>}}, \planei{\texttt{<}} and \planei{\texttt{\textbackslash}} are used to encode arcs in the first plane as above, and additional \planeii{\texttt{/}$^*$}, \planeii{\texttt{>}$^*$}, \planeii{\texttt{<}$^*$}, and \planeii{\texttt{\textbackslash}$^*$} are added to encode the arcs of the second plane. The decoding of the second plane is made with two additional separate stacks (keeping the linear-time complexity), so that arcs in different planes can always cross. This makes the encoding support relaxed 2-planar trees, and thus yields over 99\% coverage on a variety of tested treebanks.

We classify these encodings as unbounded because the number of possible labels is not bounded by a constant, but scales with respect to sentence length $n$ (consider, for example, that the first word on a sentence could have any number of \planei{\texttt{/}} between $1$ and $n-1$). Positional encodings are also unbounded, although their number of possible labels is $O(n)$ while in bracketing encodings it is $O(n^2)$. In spite of this theoretical drawback, unbounded bracketing encodings empirically tend to have fewer labels than positional encodings, and they have been shown to be a solid choice in many practical scenarios~\citep{munoz-ortiz-etal-2021-linearizations}.
 
\paragraph{Bounded bracketing encodings} \citet{gomez-rodriguez-etal-2023-4} define two encodings, derived from the basic and 2-planar bracketing encodings, but where the labels are vectors of a fixed number of bits. Thus, they are bounded, 
as the number of possible labels is a constant.

In the \emph{4-bit encoding}, each label $x_i$ is of the form $b_i^0 b_i^1 b_i^2 b_i^3$, where each $b_i^j$ is a bit: $b_i^0$ is true (false) if $w_i$ is a right (left) dependent, $b_i^1$ is true iff $w_i$ is the outermost right or left dependent of its parent node; and $b_i^2$ and $b_i^3$ are true iff $w_i$ has one or more left or right dependents, respectively. While this encoding is very compact, having a total of 16 labels, it shares the drawback of the basic brackets of not supporting same-direction crossing arcs.

The \emph{7-bit encoding} extends it using multiplanarity to support relaxed 2-planar trees by using 7 bits to represent two planes of arcs. Labels are of the form $x_i = b_i^0 \cdots b_i^6$, three more bits than the previous encoding: a bit is added to specify whether $w_i$ is a dependent in the first or second plane, and the two bits  indicating left or right dependents are split into two bits to represent the presence of such dependents in the first plane and two for the second plane. The rest of the bits retain their meaning. This encoding consistently outperformed unbounded bracketings in the experiments of~\citep{gomez-rodriguez-etal-2023-4}.

The decoding of bounded bracketing encodings back to a tree is described in detail in~\citep{gomez-rodriguez-etal-2023-4}. It works similarly to that of unbounded brackets and is also linear time, but one stack element can generate several outgoing arcs. For example, for right arcs in the first plane, we traverse the sentence from left to right. When we find a word $w_i$ with $b_i^3 = 1$ ($w_i$ has right dependents) we push \planei{\texttt{/}} to the stack. If we then find a word $w_j$ with $b_j^0 = 1$ ($w_j$ is a right dependent), we create the arc $w_i \rightarrow w_j$, but only pop the \planei{\texttt{/}} symbol from the stack if $b_j^1 = 1$ ($w_j$ is the outermost dependent). The decoding for left arcs in the first plane is analogous, but in a separate pass from right to left. For the 7-bit encoding, we add two extra passes for left and right arcs in the second plane.

\paragraph{Transition-based encodings} \citet{gomez-rodriguez-etal-2020-unifying} show that many transition-based parsers can yield sequence labeling encodings. Though in theory applicable to dependency graph parsing, previous results on syntactic parsing show that the systems' accuracy degrades for non-projective trees, so we will discard this approach.

\paragraph{Hexatagging}~\citep{amini-etal-2023-hexatagging} is the overall best-performing encoding known so far for dependency parsing. It is bounded and the most compact, as it represents projective trees with only 8 possible labels per word. Yet, its design makes it unlikely that an extension to graph parsing is possible, as it is based on projectivity and treeness (requiring converting dependency trees to a constituent-like representation). Thus, we will not use it here.

\section{Unbounded graph encodings}
Let $w = [ w_1, w_2, \ldots, w_n] \in V^n$ be a sentence. A dependency graph for $w$ is a labeled, directed graph $G=(V_w, E)$ where $V_w = \{ w_1, \ldots, w_n\}$. Contrary to dependency trees, dependency graphs in general allow reentrancy (two or more incoming arcs to the same node) and cycles. Let $G_n$ be the set of dependency graphs for sentences of length $n$. A \emph{sequence labeling encoding for dependency graph parsing} is one where $S_n=G_n$. We next present our unbounded encodings. As in dependency parsing encodings, we assume that each label $l_i$ is of the form $(d_i,x_i)$, but in this case $d_i$ is a tuple composed of the labels of the dependencies going to $w_i$, sorted by head position. As before, we focus on the encoding-dependent component $x_i$, and ignore dependency labels from now on.

\begin{figure}
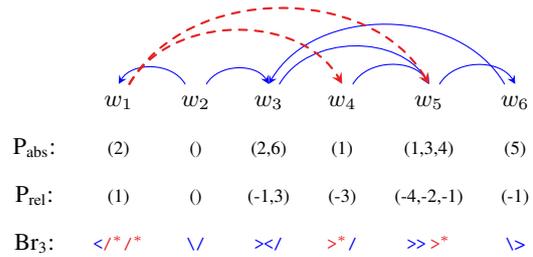
\centering\footnotesize 
    \begin{dependency}[arc edge, hide label]
        \begin{deptext}[column sep=10pt]
        \&  $w_1$ \& $w_2$ \& $w_3$ \& $w_4$ \& $w_5$ \& $w_6$ \\[1em]
        {\footnotesize P\textsubscript{abs}:} \& \scriptsize(2) \& \scriptsize() \& \scriptsize(2,6) \& \scriptsize(1) \& \scriptsize(1,3,4) \& \scriptsize(5) \\[1em]
        {\footnotesize P\textsubscript{rel}:} \& \scriptsize(1) \& \scriptsize() \& \scriptsize(-1,3) \& \scriptsize(-3) \&\scriptsize (-4,-2,-1) \& \scriptsize(-1)  \\[1em]
        {\footnotesize Br\textsubscript{3}:} 
            \&\scriptsize \planei{\texttt{<}}\planeii{\texttt{/}$^*$\texttt{/}$^*$}
            \&\scriptsize \planei{\texttt{\textbackslash/}}
            \&\scriptsize \planei{\texttt{></}}
            \&\scriptsize \planeii{\texttt{>}$^*$}\planei{\texttt{/}}
            \&\scriptsize \planei{\texttt{>}\texttt{>}} \planeii{\texttt{>}$^{*}$}
            \&\scriptsize \planei{\texttt{\textbackslash}>} \\ 
        \end{deptext}
        \depedge[edge style={plane1color}]{3}{2}{}
        \depedge[edge style={plane1color}]{3}{4}{}
        \depedge[edge style={plane1color}]{4}{6}{}
        \depedge[edge style={plane1color}]{5}{6}{}
        \depedge[edge style={plane1color}]{7}{4}{}
        \depedge[edge style={plane1color}]{6}{7}{}
        \depedge[edge style={plane2color}, thick, dashed]{2}{6}{}
        \depedge[edge style={plane2color}, thick, dashed]{2}{5}{}
    \end{dependency}
    \caption{\label{fig:example1}An example of a relaxed $2$-planar dependency graph linearized with our unbounded encodings.}
\end{figure}

\paragraph{Positional graph encodings} A naive approach for dependency graph parsing as sequence labeling is to adapt the positional encodings for dependency tree parsing. This can be done by defining $x_i$ as a tuple of arbitrary length containing the absolute (or relative) positions of all incoming arcs of each word $w_i$, so $x_i$ is an ascendingly sorted tuple with the elements of $\{h: (w_h, w_i)\in E\}$ for the naive encoding and $\{h-i: (w_h, w_i)\in E\}$ for the relative encoding. For example, in the graph in Figure~\ref{fig:example1}, $w_3$ has incoming arcs from $w_2$ and $w_6$, so the naive encoding assigns it the tuple (2,6) and the relative encoding (-1,3). Note that our definition of a dependency graph $G_n$ allows nodes with no incoming arcs, thus $x_i$ might be an empty tuple. 

\paragraph{Unbounded bracketing encodings} The extension of unbounded bracketing encodings for syntactic dependency parsing
to dependency graphs is straightforward. In this family of linearizations, the restriction to a single parent per node for syntactic parsing is achieved by explicitly enforcing exactly one incoming symbol, \planei{\texttt{<}} or \planei{\texttt{>}}, in each $x_i$. For graph parsing, we remove this restriction and allow more than one such symbol, as well as zero (which can even produce an empty string for disconnected nodes). The decoding process does not change, using
two different stacks for right and left arcs, and keeping the linear complexity. Multiplanarity is supported by introducing new sets of brackets. In Figure~\ref{fig:example1}, $w_3$ has one incoming arc from each direction (\planei{\texttt{><}}) and an outgoing arc to the right (\planei{\texttt{/}}). $w_5$ has three incoming arcs from the left, of which one is in the second plane (\planeii{\texttt{>}$^{*}$}).   

Having $k$ sets of brackets provides coverage over relaxed k-planar graphs, like the tree encoding did for relaxed 2-planar trees, and could do for relaxed k-planar trees if more sets of brackets were added. However, it is worth noting that previous work using this encoding for syntactic parsing has never experimentally explored beyond $k=2$ (i.e., adding one extra set of brackets {\planeii{\texttt{<}$^*$}, \planeii{\texttt{>}$^*$}, \planeii{\texttt{/}$^*$}, \planeii{\texttt{\textbackslash}$^*$}}). The rationale was the trend that most syntactic trees are 2-planar~\citep{gomez-rodriguez-nivre-2013-divisible}, so complicating parsing algorithms (or encodings) did not seem worthwhile for a tiny increase in coverage. However, since dependency graphs can be denser than syntactic trees, and have less propensity to be 2-planar, we also experiment with adding a third plane with a third set of brackets {\planeiii{\texttt{<}$^{**}$}, \planeiii{\texttt{>}$^{**}$}, \planeiii{\texttt{/}$^{**}$}, \planeiii{\texttt{\textbackslash}$^{**}$}}.
This gives mixed results for this particular encoding, although as will be seen later, setting $k>2$ will prove very useful to increase accuracy in the case of bounded encodings, especially with datasets containing denser graphs.

As the plane assignment algorithm (i.e. to split arcs in gold graphs into planes in a canonical way) we extend the greedy plane assignment algorithm of~\citep{strzyz-etal-2020-bracketing} to support more than two planes: we traverse arcs in order and assign each to the lowest possible plane such that it does not cross any arcs already assigned to the same plane.

\section{Bounded graph encodings}

We now define two bounded encodings for graph parsing, based on the 4- and 7-bit encodings by \citet{gomez-rodriguez-etal-2023-4}.

\subsection{$4k$-bit encoding}\label{subsec:4kbit-encoding}

\paragraph{Assumption} This encoding assumes that the set of edges $E$ of $G$ can be split into $k$ relaxed 1-planar subgraphs, such that in each subgraph, all nodes have at most one incoming arc (maximum in-degree 1). To do so, it explicitly arranges a dummy node $w_0$ that has dependencies towards any parentless node. Thus, all nodes in the graph ($w_1 \ldots w_n$) can be seen as having exactly one incoming arc (from a regular or the dummy node).

\paragraph{Encoding} It uses a sequence  of $4$ bits to encode the arcs related to the word $w_i$ that are in the $j$th subgraph. Each label $x_i$ is a grouped sequence of $4k$ bits where the $j$th group of four bits encodes only the arcs of the $j$th subgraph. The meaning of each of the four bits in a group is as in the 4-bit encoding of~\citet{gomez-rodriguez-etal-2023-4}: $b_i^{4j-4}$ is true (false) if $w_i$ has a left (right) parent in the $j$th subgraph (which could be the dummy node), $b_i^{4j-3}$ is true if $w_i$ is the farthest dependent of its parent in the $j$th subgraph, $b_i^{4j-2}$ is true if $w_i$ has left dependents and $b_i^{4j-1}$ is true if $w_i$ has right dependents in the $j$th subgraph. Thus, this encoding concatenates $k$ instances of said 4-bit encoding, which is injective and has coverage over relaxed 1-planar graphs with no more than one parent per node.\footnote{The original 4-bit encoding is described as having coverage over relaxed 1-planar forests. This is because the task, about tree parsing, forbids cycles. The encoding itself does support graphs with cycles as long as there is no reentrancy.}

\begin{figure}[tbp]\centering\footnotesize 
    \input{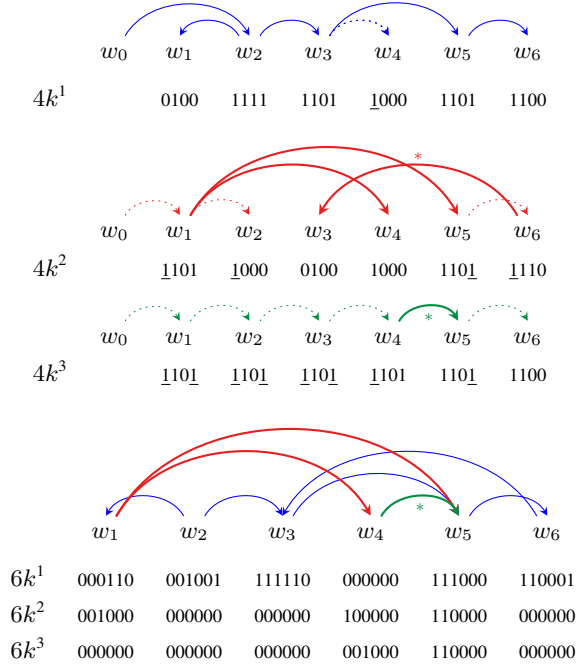}
    \vspace{-1.5em}
    \caption{\label{fig:example2}Bounded encodings for the example of Figure \ref{fig:example1}. The \emph{relaxed} 1-planar subgraphs for the $4k$-bit-encoding are shown with their linearization, added null arcs are drawn with dotted lines, and their associated bits underlined. For $6k$-bit, we use colors to distinguish the subgraph pairs. Note that, in both cases, those arcs that are assigned to different planes w.r.t. the unbounded bracketing encoding (Figure \ref{fig:example1}) are marked with $\ast$.}
\end{figure}

\paragraph{Plane assignment} We need a way to express a dependency graph as the union of $k$ relaxed 1-planar subgraphs with at most one parent per node. The plane assignment algorithm used in the unbounded bracketing does not suffice for two reasons. First (1), arcs may need to be assigned to different subgraphs not only because they cross, but also because they have the same dependent.\footnote{$4k$-bit allows only one incoming arc per node and subgraph, whose direction is encoded in the first bit of each group.} Secondly (2), nodes that have a parent in the dependency graph may be parentless in one or more subgraphs. While this may not seem problematic because the encoding supports such nodes by linking them as children of the dummy node, this would require adding arcs that can break relaxed 1-planarity. To solve (1), we modify the plane assignment algorithm to consider two arcs incompatible if they cross or share the dependent. To solve (2), we 
add artificial arcs (which we call null arcs) linking each parentless node to the immediately previous node (such an arc is guaranteed to not produce a crossing). When implementing the parser, null arcs are especially labeled and excluded from the final parse. Figure~\ref{fig:example2} shows this assignment process: note the null arcs drawn with dotted lines, and the arc $(w_4,w_5)$ being assigned to the third (green) subgraph despite not crossing any other arc, since there are already arcs going to $w_5$ in the other two subgraphs.

\paragraph{Coverage} The $4k$-bit encoding has coverage over the set of dependency graphs that can be expressed as the union of $k$ relaxed 1-planar graphs with maximum in-degree 1. This set is trivially a subset of (1) relaxed $k$-planar graphs, and (2) dependency graphs with maximum in-degree $k$. In practice, we show that $k=2$ suffices for almost total coverage on enhanced UD datasets (Table~\ref{tab:iwpt-results}), whereas on semantic dependency parsing datasets we need $k=4$ for really high coverage in most cases (Table~\ref{tab:dag-id-results}).

\paragraph{Decoding} It is done with the same linear-time algorithm as for bounded dependency parsing encodings (Section~\ref{sec:background}), with one pass per group of bits (i.e. subgraph) and arc direction, i.e., $2k$ passes. 

\subsection{$6k$-bit encoding}
\paragraph{Assumption} This encoding assumes that the parse's edge set $E$ can be split into $2k$ relaxed 1-planar subgraphs, under two conditions. First, $k$ subgraphs have all of their arcs pointing to the left, while in the other $k$, all arcs are rightward. We consider that the $2k$ subgraphs are arranged in pairs, such that each pair has a leftward subgraph and a rightward subgraph. Second, all subgraphs have maximum in-degree 1. $6k$-bit does not require dummy arcs, contrary to the $4k$-bit encoding.

\paragraph{Encoding} 
Each label $x_i$ has $k$ groups of 6 bits each. The $j$th group in each label encodes information about the $j$th pair of subgraphs (composed of the $j$th rightward subgraph and the corresponding $j$th leftward subgraph). In particular, the meaning of each of the bits in the $j$th group is as follows: $b_i^{6j-6}$ is true if $w_i$ has a parent in the $j$th rightward subgraph, $b_i^{6j-5}$ is true if $w_i$ is the farthest dependent of its parent in said subgraph, and $b_i^{6j-4}$ is true if $w_i$ has at least one dependent in said subgraph. Finally, $b_i^{6j-3}$, $b_i^{6j-2}$ and $b_i^{6j-1}$ have the same meaning for the leftward subgraph.

\paragraph{Plane assignment} To perform the assignment of arcs to subgraphs, left arcs and right arcs are processed separately (with the goal of choosing the pair whose leftward or rightward subgraph we need to assign them to). For each of these subsets, we run a modification of the plane assignment algorithm of the unbounded bracketing. Two arcs are incompatible if they cross or share the dependent (for the same reason as in the $4k$-bit encoding: subgraphs need to have maximum in-degree 1). However, here we do not need to add any null arcs, because this encoding supports representing parentless nodes natively: it suffices to set $b_i^{6j-6}$ or $b_i^{6j-3}$ to indicate that $w_i$ is parentless in the $j$th rightward (resp. leftward) subgraph. Figure~\ref{fig:example2} contains an example of this assignment, with the same graph as in previous examples. Colors depict subgraph pairs (with individual subgraphs being the subsets of leftward and rightward arcs of each color). The assignment of pairs is different from the assignment of subgraphs for $4k$-bit: apart from not needing null arcs, the arc from $w_3$ to $w_6$ can now be assigned to the first subgraph pair (vs. the second subgraph in $4k$-bit) since arcs in different directions that share a dependent can coexist in the different subgraphs of a same pair.

\paragraph{Coverage} $6k$-bit covers the set of graphs that can be split into $2k$ subgraphs meeting the above conditions. This is a subset of (1) relaxed k-planar graphs (each subgraph is relaxed 1-planar, so each subgraph pair is also relaxed 1-planar since joining a leftward graph with a rightward graph cannot generate crossings of arcs in the same direction, and the graph is the union of $k$ such pairs); and (2) graphs with maximum in-degree $2k$, since each of the $2k$ subgraphs can contribute one parent to a given node. The encoding cannot cover graphs where a node has more than $k$ incoming arcs from the same direction -- even if the in-degree does not surpass $2k$ -- as we only have $k$ rightward (leftward) subgraphs. Still, as can be seen in Tables~\ref{tab:dag-id-results} and~\ref{tab:iwpt-results}, the coverage is larger than that of the more compact $4k$-bit encoding for the same value of $k$.

\paragraph{Decoding} Again done with the same linear-time algorithm as for bounded dependency tree encodings (Section~\ref{sec:background}), with $2k$ passes, one per subgraph.

\section{Model architecture}\label{sec:model-architecture}

Let $w$ be our input sentence. The model produces a sequence of vectors $\mathbf{W}=[\mathbf{w}_1, \mathbf{w}_2,\dots,\mathbf{w}_n]$ using a generic encoder $E_\theta$. This encoder can range from lookup tables mapping words to static embeddings\footnote{However, previous work showed that continuous, contextualized representations are needed for accurate outputs.} to encoders that deeply contextualize words. We then use a generic decoder, $D_\phi$, to make output predictions at the word level. The decoder could also vary widely, from simple feed-forward networks to more sophisticated architectures. Let $\mathbf{X}=[\mathbf{x}_1,\dots \mathbf{x}_n]=D_\phi(\mathbf{W})$ be the output representations for each word. We use these outputs to predict each component $x_i$  from the label $l_i = (d_i, x_i)$. The sequence $(x_1,...,x_n)$ is fed to the corresponding encoding-specific decoding algorithm to recover the set of predicted arcs $\hat{E}$. Then, to predict the label representing the relationship between the pair of nodes of a predicted arc $(w_h,w_i)\in \hat{E}$, we concatenate their representations $[(\mathbf{w}_h | \mathbf{w}_i): (w_h, w_i) \in \hat{E}]$ and use them to feed another generic decoder $D_\varphi$ that predicts the dependency type (an element of $d_i$) associated with them.

\begin{table*}[!]
    \centering\small
    \setlength{\tabcolsep}{3pt}
    \renewcommand{\arraystretch}{1.1}
\begin{tabular}{|c|cc|c|cc|c|cc|c|cc|c|cc|c|}
    \hline 
                        & \multicolumn{3}{c|}{\bf DM\textsubscript{en}}                         &  \multicolumn{3}{c|}{\bf PAS\textsubscript{en}}                       & \multicolumn{3}{c|}{\bf PSD\textsubscript{en}}                        & \multicolumn{3}{c|}{\bf PSD\textsubscript{cs}}                        & \multicolumn{3}{c|}{\bf PAS\textsubscript{zh}}                        \\
    \cline{2-16} 
                        & \makecell[c]{\bf UF}  & \makecell[c]{\bf LF}  & \makecell[c]{\bf OF}  & \makecell[c]{\bf UF}  & \makecell[c]{\bf LF}  & \makecell[c]{\bf OF}  & \makecell[c]{\bf UF}  & \makecell[c]{\bf LF}  & \makecell[c]{\bf OF}  & \makecell[c]{\bf UF}  & \makecell[c]{\bf LF}  & \makecell[c]{\bf OF}  & \makecell[c]{\bf UF}  & \makecell[c]{\bf LF}  & \makecell[c]{\bf OF}  \\
    \hline 
    A                   & 88.66                 & 87.94                 & 100                   & 86.66                 & 85.29                 & 100                   & 89.56                 & 79.19                 & 100                  & 90.04                  & 85.33                 & 100                   & 77.12                 & 74.63                 & 100                   \\
    R                   & 91.92                 & 91.23                 & 100                   & 90.29                 & 88.86                 & 100                   & 89.74                 & 79.39                 & 100                  & 89.60                  & 84.92                 & 100                   & 79.26                 & 76.96                 & 100                   \\
    B\textsubscript{2}  & 95.16                 & 94.46                 & 99.94                 & \best{95.82}          & \best{94.31}          & 99.98                 & 92.31                 & 81.80                 & 99.83                & 92.75                  & 88.14                 & 99.83                 & 87.66                 & 85.37                 & 99.98                 \\
    B\textsubscript{3}  & 94.63                 & 93.75                 & 100                   & 95.73                 & 94.21                 & 100                   & 92.33                 & 81.65                 & 99.99                & 92.74                  & 88.02                 & 100                   & \best{87.73}          & \best{85.42}          & 100                   \\
    B4\textsubscript{2} & 86.45                 & 85.84                 & 91.23                 & 79.84                 & 78.82                 & 83.18                 & \best{92.87}          & 81.96                 & 99.68                & 92.88                  & 88.24                 & 99.65                 & 77.53                 & 75.47                 & 86.01                 \\
    B4\textsubscript{3} & 92.64                 & 91.64                 & 97.96	                & 89.65                 & 88.23                 & 93.40                 & 92.68                 & 81.99                 & 99.96                & 93.11                  & 88.33                 & 99.98                 & 83.60                 & 81.34                 & 93.42                 \\
    B4\textsubscript{4} & 95.07                 & 94.35                 & 99.64                 & 93.79                 & 92.35                 & 97.59                 & 92.80                 & 82.00                 & 100                  & \best{93.39}           & \best{88.79}          & 100                   & 81.40                 & 78.68                 & 96.60                 \\
    B6\textsubscript{2} & 91.21                 & 90.67                 & 96.37                 & 87.70                 & 86.64                 & 91.59                 & 92.66                 & 81.88                 & 99.77                & 93.37                  & 88.54                 & 99.79                 & 81.88                 & 79.81                 & 92.09                 \\
    B6\textsubscript{3} & 94.90                 & 94.15                 & 99.51                 & 93.58                 & 92.16                 & 97.56                 & 92.61                 & \best{82.13}          & 99.98                & 93.44                  & 88.61                 & 99.98                 & 85.67                 & 83.50                 & 96.57                 \\
    B6\textsubscript{4} & \best{95.23}          & \best{94.52}          & 99.96	                & 95.32                 & 93.87                 & 99.37                 & 92.74                 & 81.89                 & 100                  & 93.30                  & 88.45                 & 100                   & 87.06                 & 84.77                 & 98.30                 \\
    \hline 
    Biaf                & 95.07                 & 94.31                 & 100                   & 95.69                 & 94.12                 & 100                   & 92.95                 & 82.08                 & 100                  & 93.65                  & 88.73                 & 100                   & 87.80                 & 85.49                 & 100                   \\ 
    \hline 
\end{tabular}
    \caption{\label{tab:dag-id-results} DAG in-distribution results. OF is the coverage of each encoding in terms of oracle F-score. }
\end{table*}

\section{Experiments}
For evaluation, we use two kinds of dependency graph structures with different properties: semantic dependency parsing, and enhanced UD parsing.

\paragraph{Datasets} For semantic dependency parsing, we will use the following datasets and formalisms from the SemEval 2018 Task 18 \cite{oepen-etal-2015-semeval}: (i) the English dataset annotated with DELPH-IN MRS-Derived Bi-Lexical Dependencies \citep[DM][]{ivanova-etal-2012-contrastive} (ii), the English and Chinese datasets with Enju Predicate–Argument Structures \citep[PAS][]{miyao2005corpus}, and (iii) the English and Czech datasets with Prague Semantic Dependencies \citep[PSD][]{hajic-etal-2012-announcing}. They are all collections of sentences annotated as graphs, where some tokens do not contribute to the graph, some might have just one parent, and others multiple parents. These representations are DAGs, excluding the study of relevant phenomena involving cycles. To study cycles, we rely on enhanced universal dependencies, particularly on the version released together with the IWPT 2021 Shared Task \cite{bouma-etal-2021-raw}. We evaluated five languages: Arabic, Finnish, French, Slovak, and Tamil, joining the different treebanks of the same language to build a larger annotated corpus. In Appendix \ref{ap:stats}, we provide additional information and statistics for both DAG and IWPT treebanks.

\paragraph{Metrics} We use the SDP evaluation toolkit\footnote{\href{https://github.com/semantic-dependency-parsing/toolkit}{github.com/semantic-dependency-parsing/toolkit}} \citep{oepen-etal-2015-semeval}. We report both unlabeled and labeled F1 score (LF, UF) w.r.t. the predicted dependencies, i.e. triplets of the form (predicate, role, argument).\footnote{Detection of root nodes is considered as identifying additional virtual dependencies, and counts for evaluation.} 
Results with further metrics, including  exact match, are in the Appendix.

\paragraph{Models' setup} For the encoders of our taggers, we use a few representatives. We focus on experiments using two Transformer-based encoders: (1) XLM-RoBERTa \cite{conneau-etal-2020-unsupervised}, as a single multilingual pre-trained encoder for non-English experiments, and (2) XLNet \cite{yang2019xlnet} for English experiments. In addition, we experimented with biLSTMs for comparison in terms of speed-accuracy trade-off. The decoder is a 1-layered feed-forward network followed by a softmax. We did a minor hyperparameter search to tune the optimizer and batch size to our resources.\footnote{See Appendix \ref{ap:setup} for more details about the model size and training specifications.} For comparison, we include an external model (Biaf), a biaffine semantic dependency parser \cite{dozat-manning-2018-simpler} - from \texttt{supar}\footnote{\href{https://github.com/yzhangcs/parser}{github.com/yzhangcs/parser}} - using the same encoders as for our experiments.

\subsection{Empirical results}

We provide here the main results of our experiments. Supplementary data and insights can be found in Appendix \ref{ap:results}.

For brevity, we refer to absolute and relative graph positional encodings as A and R. Unbounded bracketing encodings are denoted as B$_k$, where $k$ is the number of supported planes. For $4k$-bit and $6k$-bit encodings, we use B4$_k$ and B6$_k$, respectively, where the subindex is the value of $k$ (number of subgraphs or pairs the graph is divided into).

\begin{table*}[tbp]
    \centering\small
    \setlength{\tabcolsep}{4pt}
    \renewcommand{\arraystretch}{1.1}
    \begin{tabular}{|c|cc|c|cc|c|cc|c|cc|c|cc|c|}
    \hline 
                        & \multicolumn{3}{c|}{\bf ar}                   & \multicolumn{3}{c|}{\bf fi}                   &  \multicolumn{3}{c|}{\bf fr}                  & \multicolumn{3}{c|}{\bf sk}                   & \multicolumn{3}{c|}{\bf ta}                   \\
    \cline{2-16} 
                        & \textbf{UF}   & \textbf{LF}   & \textbf{OF}   & \textbf{UF}   & \textbf{LF}   & \textbf{OF}   & \textbf{UF}   & \textbf{LF}   & \textbf{OF}   & \textbf{UF}   & \textbf{LF}   & \textbf{OF}   & \textbf{UF}   & \textbf{LF}   & \textbf{OF}   \\
    \hline 
    A                   & 75.00         & 69.17         & 100           & 84.52         & 80.97         & 100           & 80.79         & 76.58         & 100           & 83.53         & 79.39         & 100           & 34.22         & 27.27         & 100           \\
    R                   & 82.06         & 75.58         & 100           & 85.90         & 82.34         & 100           & 83.06         & 78.66         & 100           & 86.79         & 82.59         & 100           & 64.69         & 53.69         & 100           \\
    B\textsubscript{2}  & 87.85         & 80.98         & 99.82         & 91.19         & 88.16         & 99.60         & 91.06         & 87.60         & 99.97         & 93.31         & 90.15         & 99.79         & 73.62         & 62.02         & 100           \\
    B\textsubscript{3}  & 87.82         & 81.22         & 99.94         & 91.20         & 88.13         & 99.94         & 90.59         & 86.65         & 100           & 93.81         & 90.44         & 99.96         & 73.62         & 62.02         & 100           \\
    B4\textsubscript{2} & 87.84         & 81.21         & 99.77         & 91.53         & 88.46         & 99.60         & 92.37         & 88.64         & 99.87         & 93.79         & 90.41         & 99.72         & 75.81         & 63.36         & 99.95         \\
    B4\textsubscript{3} & 87.81         & 81.11         & 99.90         & 91.58         & 88.41         & 99.87         & 92.60         & 88.88         & 99.98         & 94.08         & 90.51         & 99.94         & 76.01         & \best{64.54}  & 100           \\
    B4\textsubscript{4} & 88.09         & 81.27         & 99.94         & 91.64         & 88.56         & 99.94         & 92.61         & 88.47         & 100           & 93.85         & 90.32         & 99.99         & 76.01         & 64.54         & 100           \\
    B6\textsubscript{2} & 87.68         & 80.97         & 99.85         & 92.02         & 89.03         & 99.69         & 92.60         & 88.82         & 99.93         & 94.06         & 90.86         & 99.85         & \best{76.16}  & 63.48         & 100           \\
    B6\textsubscript{3} & 88.10         & 81.37         & 99.93         & 91.64         & 88.71         & 99.89         & 91.66         & 88.06         & 99.99         & \best{94.26}  & \best{90.94}  & 99.97         & 76.16         & 63.48         & 100           \\
    B6\textsubscript{4} & \best{88.33}  & \best{81.60}  & 99.95         & \best{92.03}  & \best{89.07}  & 99.95         & \best{92.96}  & \best{89.80}  & 100           & 93.89         & 90.73         & 99.99         & 76.16         & 63.48         & 100           \\
    \hline 
    Biaf                & 89.70         & 82.48         & 100           & 93.54         & 90.93         & 100           & 93.71         & 89.77         & 100           & 94.22         & 90.61         & 100           & 76.08         & 65.74         & 100           \\ 
    \hline 
\end{tabular}

    \caption{\label{tab:iwpt-results} EUD parsing results on IWPT datasets. Notation as in Table~\ref{tab:dag-id-results}.}
\end{table*}

Table~\ref{tab:dag-id-results} shows the results for the DAG experiments, including DM, PAS, and PSD in-domain test data, for all our encodings. The results on the out-of-domain test and development sets are in the Appendix (Table~\ref{tab:dag-ood-results}), but the trends are similar.

Positional encodings yield, overall, less accurate results for dependency graph parsing, possibly influenced by the added complexity of representing a list of head positions, rather than just one as in syntactic parsing, into a single label. The rest of our encodings, though, are much more robust: most of them are rougly on par or even outperform the biaffine parser, a competitive baseline. The relative performance of different bracketings strongly depends on whether bounded encodings obtain high coverage. In the English and Czech PSD datasets, which are comparatively less dense, bounded encodings achieve almost total theoretical coverage (column OF on the table) and excel in performance, and the more compact $4k$-bit encodings seem to be slightly better than $6k$-bit encodings. However, in PAS datasets, where the coverage of bounded encodings is lower due to higher graph density (see Appendix Table~\ref{tab:stats} for graph density statistics), unbounded brackets clearly outperform bounded brackets, and $4k$-bit encodings suffer more than $6k$-bit ones, due to having even less coverage. The DM dataset is a middle ground, both in terms of coverage and results, with unbounded, B4$_4$ and B6$_4$ obtaining fairly similar accuracies.

With respect to the parameter $k$ (i.e., number of subgraphs or pairs used in the bracketing encodings), it is important to increase it beyond $2$ for bounded encodings in the denser datasets, where coverage and accuracy increase hand in hand when setting $k$ to $3$ and $4$. This differs w.r.t. common practice in encodings for dependency tree parsing, where values of $k$ beyond $2$ are not used in experiments due to $k=2$ providing almost full coverage. The case for $k>2$ is less clear for unbounded bracketing, where $k=2$ already provides fairly good coverage. $B_3$ is still better than $B_2$ in one dataset,
but this might be due to statistical noise.

For EUD parsing (Table~\ref{tab:iwpt-results}), in spite of the presence of cycles, graphs are sparser than in our DAG datasets, so all encodings have fairly good coverage. In this context, bounded encodings obtain the best performance, consistent with the DAG results, although in this case they fall somewhat short of the biaffine parser for some languages, being on par in others. $6k$-bit encodings generally outperform $4k$-bit encodings. The $4k$-bit encoding performs better (at least in terms of labelled F-score) for Tamil, maybe due to the much smaller training set favoring compact encodings.  The absolute and relative encodings (prone to sparsity problems) obtain even poorer results on Tamil.

Finally, with respect to the parameter $k$, the effect of increasing it beyond $2$ in the EUD datasets is not as clear as in denser datasets, which makes sense as the increases in coverage afforded by larger values of $k$ is minimal. However, there seems to be a clearly positive (if weak) influence in the case of bounded encodings (where the best results are always achieved by encodings with $k$ at least $3$, except in the case of Tamil, where the value of $k$ mostly does not matter as all but one encoding reach 100\% coverage). This suggests that increasing $k$ might be useful even when the coverage increase is small. On the other hand, for unbounded bracketings, no clear trend is visible, with $B_3$ outperforming $B_2$ in some datasets and underperforming it in others.

\subsection{Speed analysis}\label{subsec:speed}

\begin{figure}[hbtp!]
    \begin{subfigure}{0.48\textwidth}
        \caption{DM (English) in-distribution set.}
        \includegraphics[width=\textwidth]{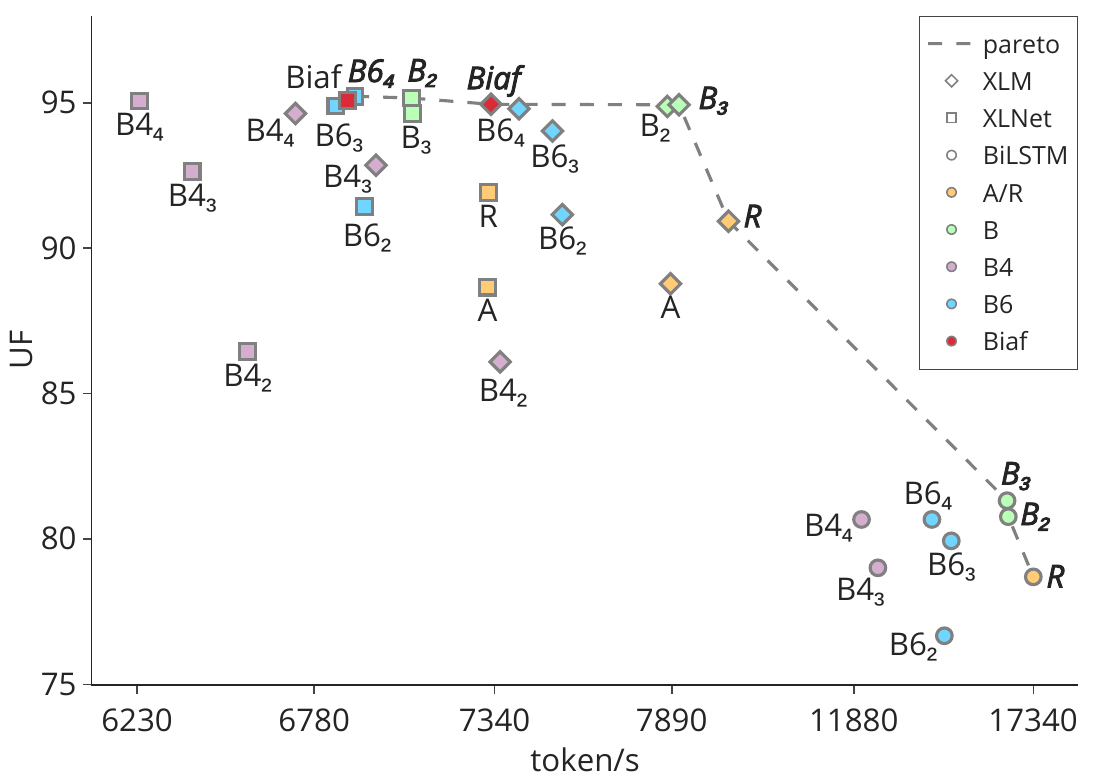}
    \end{subfigure}
    \begin{subfigure}{0.48\textwidth}
        \caption{\label{fig:pareto-ar}Arabic-PADT test set.}
        \includegraphics[width=\textwidth]{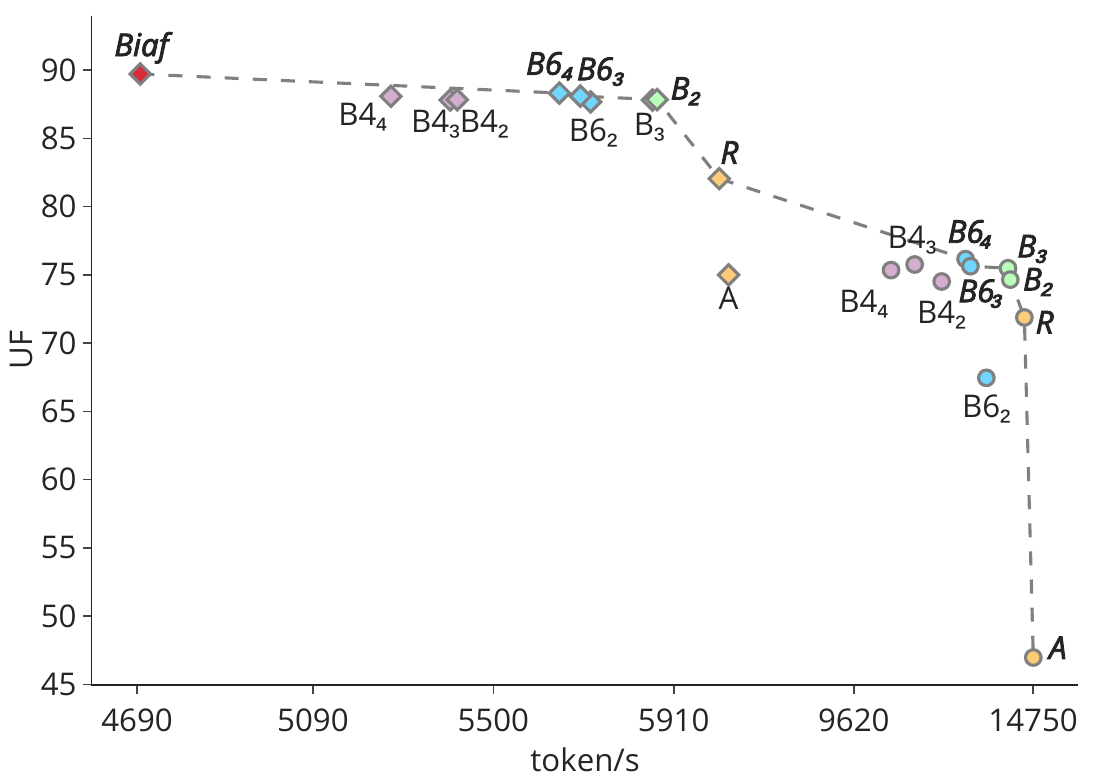}
    \end{subfigure}
    \begin{subfigure}{0.48\textwidth}
        \caption{French-Sequoia/FQB test set.}
        \includegraphics[width=\textwidth]{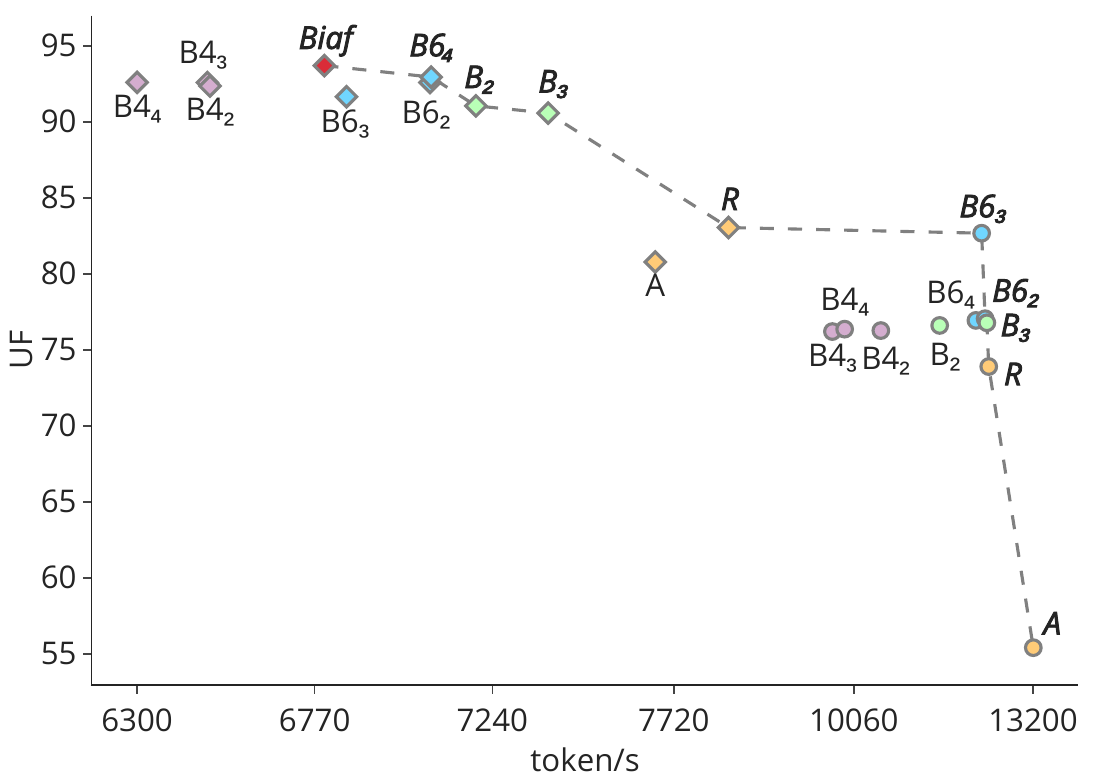}
    \end{subfigure}
    \caption{\label{fig:pareto}Pareto front: UF vs. speed. X-axis rescaled, outliers omitted for clarity.}
\end{figure}

In Figure \ref{fig:pareto}, we present a Pareto front in terms of UF and speed (tokens per second) across three datasets: DM (English, in-distribution) and two IWPT corpora (Arabic and French). In Appendix \ref{ap:results} we include additional figures for other treebanks to provide a more complete picture.

Alongside the biaffine baseline, we include a lighter encoder, a two-stacked biLSTM. Note that, for space reasons, we did not include biLSTM results in the previous section, as the UF performance is substantially lower, as seen in the figure. The 2-layered BiLSTM is faster than the pretrained encoders in all cases, followed by XLM-R and XLNet. Although BiLSTMs are less accurate, they can still appear on the Pareto front in some setups.

Overall, our parsers offer either similar or faster speeds compared to the biaffine model, with the exception of those relying on the $4k$-bit encoding, which consistently lag behind the rest of the models across all datasets. This may be due to the extra dependencies generated in the encoding step (Section \ref{subsec:4kbit-encoding}), which force the decoder to process more arcs in the decoding step and predict more dependency types. The positional encodings (A, R), instead, offer the best inference speed.\footnote{Our decoding process is parallel for positional encodings, since each label $x_i$ can be independently processed to create a subset of predicted arcs, thus optimizing the inference time.}

\section{Conclusion}
For the first time, we have framed dependency graph parsing tasks, like semantic dependency or EUD parsing, as sequence labeling tasks. We have proposed a wide variety of bounded and unbounded encodings that -- with the right representation -- can be learned by standard encoders. Among unbounded encodings, positional strategies performed poorly, but bracketing-based encoders obtained robust performance, excelling especially in dense datasets. On the other hand, the more compact bounded encodings, with a fixed number of bits per label, obtained the best results in sparser datasets. Overall, results are comparable or even outperform a strong biaffine baseline. Thus, dependency graph parsing can effectively be solved as sequence labeling, as both bounded and unbounded encodings are learnable using standard bidirectional encoders and simple feed-forward decoders.

\section*{Limitations}

\paragraph{Anchoring} 
This work focuses on dependency graph parsing, i.e. structured prediction problems where the input is a sentence and the output is a graph where nodes correspond to words. While this template fits a considerable range of tasks, including several flavors of semantic dependency parsing, EUD parsing or graph-based sentiment analysis; there are kinds of meaning representation parsing that do not fit this framework. More in detail, meaning representations can be hierarchically organized in different formal flavors for semantic graphs, as described by \citet{oepen-etal-2019-mrp, oepen-etal-2020-mrp}. These \emph{flavors} refer to the relationship between the words of a sentence and the nodes of the graph (known as \emph{anchoring}). The scope of our work includes flavor (0) representations. In this flavor, the nodes of the graphs are the tokens of the input sentence, meaning there is a one-to-one correspondence between the nodes of the graph and the words. In future work, we aim to generalize our encodings to more relaxed flavors. However, we would like to remark that the technical contributions to casting these flavors as sequence labeling would mainly arise from aspects other than the linearizations of the graphs. Once the nodes of such graphs are computed, our linearizations could be directly applied to any dependency-based formalism.

\paragraph{Physical resources} Our computational resources are limited to eight shared RTX 3090 GPUs. Despite this, we have successfully trained models for various formalisms and languages. This has given us insight into the learnability of different encodings. While more powerful architectures might boost our results, our extensive empirical findings clearly support the key contributions of our work.

\section*{Ethical considerations}

We do not observe ethical implications in our work. Our research focuses on improving the technical aspects of semantic dependency parsing and enhanced Universal Dependencies parsing, which are primarily computational and linguistic challenges. The methodologies and applications discussed do not involve sensitive personal data, human subjects, or scenarios that could lead to ethical concerns. Thus, our findings and techniques can be applied within the field without ethical reservations.

We acknowledge the environmental impact associated with the training process in neural models, particularly in terms of CO\textsubscript{2} emissions, with Spain being the country where the experiments were conducted. We measured the carbon footprint per epoch of our models at training and inference time: 0.24 g CO\textsubscript{2} (training) and 0.06 g CO\textsubscript{2} (inference) for the larger pretrained architectures, and 0.06 g CO\textsubscript{2} (training) and 0.02 g CO\textsubscript{2} (inference) for the non-pretrained architectures. These metrics are relatively low compared with the latest models in NLP research, encouraging future work to consider energy-efficient approaches. Additionally, for reference, the European Union sets a maximum of around 115 g CO\textsubscript{2} per kilometer for newly manufactured cars.

\section*{Acknowledgments}

We acknowledge the European Research Council (ERC), which has funded this research under the Horizon Europe research and innovation programme (SALSA, grant agreement No 101100615). We also acknowledge grants SCANNER-UDC (PID2020-113230RB-C21) funded by MICIU/AEI/10.13039/501100011033; GAP (PID2022-139308OA-I00) funded by MICIU/AEI/10.13039/501100011033/ and ERDF, EU; LATCHING (PID2023-147129OB-C21) funded by MICIU/AEI/10.13039/501100011033 and ERDF, EU; and TSI-100925-2023-1 funded by Ministry for Digital Transformation and Civil Service and ``NextGenerationEU'' PRTR; as well as funding by Xunta de Galicia (ED431C 2024/02), and Centro de Investigación de Galicia ``CITIC'', funded by the Xunta de Galicia through the collaboration agreement between the Consellería de Cultura, Educación, Formación Profesional e Universidades and the Galician universities for the reinforcement of the research centres of the Galician University System (CIGUS).

\bibliography{anthology,custom}

\appendix

\section{Appendix}\label{sec:appendix}

\subsection{Behavior in degenerate cases}\label{ap:denegerate-cases}
All of the encodings presented reduce to existing dependency tree encodings in degenerate cases. In the case of the positional and unbounded bracketing encoding, when using them to encode a corpus containing only trees, the result will be as with the positional and unbounded tree encodings (the former will always generate tuples of size one, which are like the labels of the corresponding tree encoding, and the latter will generate the exact same labels with at most one incoming arc symbol as the unbounded bracketing encoding for trees).

In the case of $4k$-bit bracketing, if $k=1$, the encoding reduces exactly to the existing $4$-bit encoding for projective trees. Even if one used $k>1$ for a corpus of projective trees, the result would be isomorphic, as extra planes would not be used so every group of bits beyond the first would always take the same values.

Finally, the $6k$-bit bracketing for $k=1$ also becomes isomorphic to the $4$-bit encoding: even though there are six bits, when the graph is a tree there are only four possible combinations of $b_i^0$, $b_i^1$, $b_i^3$ and $b_i^4$ (which vary depending on whether a node is a left or right dependent and whether it is the outermost dependent, with combinations that make it have two or zero parents excluded) and these are isomorphic to the four possible combinations of $b_i^0$ and $b_i^1$ in the 4-bit encoding, so there is a structure-preserving bijection between the labels generated by both encodings. The only caveat is that one needs to consider the root node as outermost right dependent of a dummy root node $w_0$.

This means that, except for the change in root node handling for the $6k$-bit encoding, there is no extra complexity added, and no possible change in accuracy,\footnote{This is guaranteed in practice because unused labels cannot harm accuracy in any way, as sequence labeling systems only train on the labels that they effectively see in the training set, not on the theoretical space of labels.} if one uses the encodings presented here to train a model on a treebank of projective trees.

\subsection{Treebank statistics}\label{ap:stats}
The treebanks released in the IWPT 2021 Shared Task and used in this work \cite{bouma-etal-2021-raw} were joined (preserving the original train, validation and test split) to train each language in a single data benchmark, as specified in Table \ref{tab:iwpt-benchmark}. In Table \ref{tab:stats}, we summarize some general statistics about the datasets used to train and evaluate our models. Table \ref{tab:enc-stats} shows a summary of the labels generated by our encodings for each dataset used.

\begin{table}
    \centering\scriptsize
    \setlength{\tabcolsep}{2.6pt}
    \renewcommand{\arraystretch}{1.1}
    \begin{tabular}{c|c|c|c}
        \hline 
        &  \textbf{train} & \textbf{dev} & \textbf{test}\\
        \hline 
        \textbf{ar} & PADT & PADT & PADT \\ 
        \textbf{fi} & TDT & TDT & TDT, PUD \\ 
        \textbf{fr} & Sequoia & Sequoia &  Sequoia, FQB \\ 
        \textbf{sk} & SNK & SNK & SNK\\ 
        \textbf{ta} & TTB & TTB & TTB \\ 
        \hline 
    \end{tabular}
    \caption{\label{tab:iwpt-benchmark}Multilingual benchmark obtained from joining the different treebanks from the IWPT 2021 datasets \cite{bouma-etal-2021-raw}.}
\end{table}
\begin{table}
    \centering\scriptsize
    \setlength{\tabcolsep}{1.7pt}
    \renewcommand{\arraystretch}{1.05}
    
\begin{tabular}{p{0.3cm}|c|ccccc|cccc|c}
    \hline
     & \multirow{2}{*}{\bf\#sents} & \multicolumn{5}{c|}{\bf\% planes} & \multirow{2}{*}{\bf h/n} & \multirow{2}{*}{\bf d/n} & \multirow{2}{*}{\bf a/g} &  \multirow{2}{*}{\bf len}& \multirow{2}{*}{\bf\#cycs.}\\
     \cline{3-7}
     & & 1 & 2 & 3 & 4 & 5 & & &  \\
    \hline
    \multicolumn{11}{c}{\bf DAG} \\
    \hline 
    \parbox[t]{2mm}{\multirow{4}{*}{\rotatebox[origin=c]{90}{\textbf{DM\textsubscript{en}}}}} 
    & 33964 & 57.44 & 41.39 & 1.16 & 0.01 & 0.00 & 0.79 & 0.75 & 17.68 & 22.52 & 0 \\
    & 1692 & 59.69 & 39.30 & 1.00 & 0.00 & 0.00 & 0.79 & 0.75 & 17.55 & 22.28 & 0 \\
    & 1410 & 53.19 & 44.96 & 1.77 & 0.07 & 0.00 & 0.78 & 0.74 & 17.60 & 22.66 & 0 \\
    & 1849 & 59.17 & 40.02 & 0.81 & 0.00 & 0.00 & 0.77 & 0.73 & 13.11 & 17.08 & 0 \\
    \hline
    \parbox[t]{2mm}{\multirow{4}{*}{\rotatebox[origin=c]{90}{\textbf{PAS\textsubscript{en}}}}} 
    & 33964 & 53.46 & 45.88 & 0.67 & 0.00 & 0.00 & 1.02 & 0.98 & 22.96 & 22.52 & 0 \\
    & 1692 & 58.81 & 40.43 & 0.77 & 0.00 & 0.00 & 1.01 & 0.97 & 22.53 & 22.28 & 0 \\
    & 1410 & 51.21 & 48.09 & 0.71 & 0.00 & 0.00 & 1.02 & 0.98 & 23.15 & 22.66 & 0 \\
    & 1849 & 56.14 & 43.16 & 0.70 & 0.00 & 0.00 & 1.02 & 0.96 & 17.35 & 17.08 & 0 \\
    \hline
    \parbox[t]{2mm}{\multirow{4}{*}{\rotatebox[origin=c]{90}{\textbf{PSD\textsubscript{en}}}}} 
    & 33964 & 58.27 & 36.02 & 5.03 & 0.61 & 0.06 & 0.70 & 0.67 & 15.80 & 22.52 & 0 \\
    & 1692 & 59.10 & 35.64 & 4.37 & 0.65 & 0.18 & 0.71 & 0.68 & 15.74 & 22.28 & 0 \\
    & 1410 & 55.89 & 38.01 & 5.39 & 0.71 & 0.00 & 0.70 & 0.67 & 15.79 & 22.66 & 0 \\
    & 1849 & 64.85 & 28.39 & 5.90 & 0.76 & 0.11 & 0.68 & 0.64 & 11.57 & 17.08 & 0 \\
    \hline
    \parbox[t]{2mm}{\multirow{4}{*}{\rotatebox[origin=c]{90}{\textbf{PSD\textsubscript{cs}}}}} 
    & 40047 & 55.67 & 37.72 & 5.86 & 0.69 & 0.06 & 0.78 & 0.75 & 18.34 & 23.45 & 0 \\
    & 2010 & 58.56 & 35.62 & 5.42 & 0.30 & 0.10 & 0.78 & 0.75 & 17.97 & 22.99 & 0 \\
    & 1670 & 55.81 & 38.32 & 5.39 & 0.48 & 0.00 & 0.77 & 0.74 & 17.71 & 22.99 & 0 \\
    & 5226 & 66.84 & 26.48 & 5.59 & 0.92 & 0.15 & 0.78 & 0.74 & 13.19 & 16.82 & 0 \\
    \hline
    \parbox[t]{2mm}{\multirow{3}{*}{\rotatebox[origin=c]{90}{\textbf{PAS\textsubscript{zh}}}}} 
    & 25896 & 65.56 & 33.51 & 0.92 & 0.00 & 0.00 & 1.02 & 0.98 & 22.95 & 22.43 & 0 \\
    & 2440 & 61.52 & 36.89 & 1.60 & 0.00 & 0.00 & 1.02 & 0.99 & 28.60 & 27.95 & 0 \\
    & 8976 & 64.43 & 34.44 & 1.14 & 0.00 & 0.00 & 1.02 & 0.98 & 24.47 & 23.89 & 0 \\
    \hline
    \multicolumn{11}{c}{\bf IWPT} \\
    \hline 
    \parbox[t]{2mm}{\multirow{3}{*}{\rotatebox[origin=c]{90}{\textbf{ar}}}}
    & 6075 & 63.80 & 31.23 & 4.13 & 0.67 & 0.15 & 1.06 & 1.04 & 39.22 & 36.85 & 1386 \\
    & 909 & 68.10 & 27.06 & 4.29 & 0.33 & 0.22 & 1.06 & 1.03 & 35.24 & 33.27 & 225 \\
    & 680 & 62.65 & 32.35 & 3.82 & 1.03 & 0.15 & 1.06 & 1.04 & 44.11 & 41.56 & 178 \\
    \hline
    \parbox[t]{2mm}{\multirow{3}{*}{\rotatebox[origin=c]{90}{\textbf{fi}}}} 
    & 12217 & 64.89 & 27.58 & 6.61 & 0.78 & 0.09 & 1.08 & 1.01 & 14.41 & 13.33 & 1855 \\
    & 1364 & 62.83 & 28.89 & 7.55 & 0.37 & 0.29 & 1.09 & 1.01 & 14.57 & 13.42 & 203 \\
    & 2555 & 68.96 & 25.28 & 5.17 & 0.51 & 0.04 & 1.06 & 0.99 & 15.24 & 14.44 & 414 \\
    \hline
    \parbox[t]{2mm}{\multirow{3}{*}{\rotatebox[origin=c]{90}{\textbf{fr}}}}
    & 2231 & 69.03 & 29.05 & 1.79 & 0.13 & 0.00 & 1.05 & 1.01 & 23.77 & 22.64 & 546 \\
    & 412 & 67.72 & 30.83 & 1.21 & 0.24 & 0.00 & 1.05 & 1.01 & 25.51 & 24.28 & 112 \\
    & 2745 & 90.02 & 9.40 & 0.58 & 0.00 & 0.00 & 1.02 & 0.95 & 12.76 & 12.45 & 193 \\
    \hline
    \parbox[t]{2mm}{\multirow{3}{*}{\rotatebox[origin=c]{90}{\textbf{sk}}}}
    & 8483 & 73.00 & 25.49 & 1.44 & 0.05 & 0.02 & 1.05 & 0.95 & 9.96 & 9.50 & 469 \\
    & 1060 & 70.85 & 26.98 & 2.08 & 0.09 & 0.00 & 1.06 & 0.98 & 12.70 & 12.01 & 105 \\
    & 1061 & 67.01 & 29.88 & 2.83 & 0.19 & 0.09 & 1.07 & 0.98 & 12.79 & 12.00 & 117 \\
    \hline
    \parbox[t]{2mm}{\multirow{3}{*}{\rotatebox[origin=c]{90}{\textbf{ta}}}}
    & 400 & 89.50 & 10.50 & 0.00 & 0.00 & 0.00 & 1.02 & 0.96 & 16.19 & 15.82 & 1 \\
    & 80 & 87.50 & 12.50 & 0.00 & 0.00 & 0.00 & 1.06 & 0.99 & 16.68 & 15.79 & 22 \\
    & 120 & 91.67 & 8.33 & 0.00 & 0.00 & 0.00 & 1.04 & 0.98 & 17.29 & 16.57 & 38 \\
    \hline
\end{tabular}

    \caption{\label{tab:stats}Number of sentences (\textbf{\#sents}) and cycles  (\textbf{\#cycs.}); ratio of $\{1...5\}$-planar graphs (\textbf{\%planes}); average number of heads (\textbf{h/n}) and dependants per node (\textbf{d/n}), arcs per graph (\textbf{a/g}) and sentence length (\textbf{len}). Each subrow corresponds to the train, dev, i.d. and o.o.d (test) split of each treebank in the DAG (IWPT) dataset.}
\end{table}

\begin{table}
    \centering\scriptsize
    \setlength{\tabcolsep}{1.7pt}
    \renewcommand{\arraystretch}{1.05}
    \begin{tabular}{|c|cccccccccc|c|}
    \cline{2-12}
    \multicolumn{1}{c|}{}& \textbf{A} & \textbf{R} & \textbf{B\textsubscript{2}} & \textbf{B\textsubscript{3}} & \textbf{B4\textsubscript{2}} & \textbf{B4\textsubscript{3}} & \textbf{B4\textsubscript{4}} & \textbf{B6\textsubscript{2}} & \textbf{B6\textsubscript{3}} & \textbf{B6\textsubscript{4}} & $|\mathcal{R}|$ \\ 
    \cline{2-12}
    \multicolumn{1}{c}{}& \multicolumn{11}{|c|}{\bf DAG} \\
    \hline 
    \parbox[t]{2mm}{\multirow{4}{*}{\rotatebox[origin=c]{90}{\textbf{DM\textsubscript{en}}}}} 
        & 12666 & 7601 & 622 & 714 & 159 & 526 & 882 & 252 & 486 & 621 & 60 \\
        & 2226 & 1249 & 257 & 273 & 116 & 285 & 426 & 144 & 231 & 280 & 44 \\
        & 1959 & 1133 & 269 & 292 & 116 & 279 & 406 & 148 & 226 & 263 & 44 \\
        & 1606 & 926 & 288 & 307 & 120 & 293 & 416 & 156 & 246 & 281 & 48 \\
    \hline
    \parbox[t]{2mm}{\multirow{4}{*}{\rotatebox[origin=c]{90}{\textbf{PAS\textsubscript{en}}}}} 
        & 23664 & 17434 & 1510 & 1590 & 159 & 587 & 1254 & 212 & 543 & 907 & 43 \\
        & 3041 & 1936 & 449 & 464 & 117 & 320 & 550 & 111 & 239 & 364 & 40 \\
        & 2792 & 1842 & 476 & 488 & 116 & 320 & 561 & 128 & 263 & 401 & 42 \\
        & 2421 & 1633 & 539 & 555 & 122 & 372 & 643 & 121 & 279 & 441 & 42 \\
    \hline 
    \parbox[t]{2mm}{\multirow{4}{*}{\rotatebox[origin=c]{90}{\textbf{PSD\textsubscript{en}}}}}
        & 2424 & 2978 & 1302 & 1890 & 210 & 677 & 986 & 360 & 677 & 819 & 91 \\
        & 617 & 702 & 436 & 533 & 144 & 299 & 356 & 185 & 265 & 283 & 81 \\
        & 570 & 647 & 401 & 500 & 146 & 286 & 340 & 166 & 237 & 256 & 78 \\
        & 574 & 715 & 406 & 501 & 146 & 299 & 366 & 168 & 239 & 261 & 75 \\
    \hline 
    \parbox[t]{2mm}{\multirow{4}{*}{\rotatebox[origin=c]{90}{\textbf{PSD\textsubscript{cs}}}}}
        & 3768 & 4676 & 2155 & 3057 & 225 & 864 & 1323 & 413 & 861 & 1050 & 62 \\
        & 743 & 885 & 550 & 671 & 159 & 347 & 427 & 216 & 305 & 323 & 62 \\
        & 712 & 840 & 570 & 693 & 156 & 338 & 401 & 196 & 281 & 300 & 58 \\
        & 1107 & 1465 & 822 & 1084 & 191 & 537 & 705 & 275 & 460 & 523 & 65 \\
    \hline 
    \parbox[t]{2mm}{\multirow{3}{*}{\rotatebox[origin=c]{90}{\textbf{PAS\textsubscript{zh}}}}}
        & 26749 & 21334 & 1603 & 1747 & 158 & 567 & 1128 & 189 & 474 & 774 & 33 \\
        & 6040 & 4237 & 660 & 699 & 129 & 374 & 644 & 144 & 307 & 438 & 31 \\
        & 13557 & 10259 & 1105 & 1197 & 143 & 467 & 896 & 163 & 377 & 606 & 32 \\
    \hline 
    \multicolumn{1}{c}{}& \multicolumn{11}{|c|}{\bf IWPT} \\
    \hline
    \parbox[t]{2mm}{\multirow{3}{*}{\rotatebox[origin=c]{90}{\textbf{ar}}}} 
        & 2543 & 2368 & 872 & 1218 & 149 & 349 & 468 & 217 & 368 & 429 & 1056 \\
        & 714 & 657 & 363 & 425 & 106 & 180 & 209 & 122 & 160 & 174 & 356 \\
        & 917 & 742 & 362 & 417 & 106 & 174 & 206 & 119 & 149 & 166 & 363 \\
    \hline 
    \parbox[t]{2mm}{\multirow{3}{*}{\rotatebox[origin=c]{90}{\textbf{fi}}}} 
        & 1630 & 1593 & 1192 & 1729 & 185 & 537 & 745 & 226 & 415 & 505 & 417 \\
        & 528 & 484 & 468 & 590 & 143 & 292 & 342 & 147 & 216 & 239 & 191 \\
        & 485 & 526 & 517 & 677 & 150 & 318 & 381 & 152 & 215 & 238 & 234 \\
    \hline 
    \parbox[t]{2mm}{\multirow{3}{*}{\rotatebox[origin=c]{90}{\textbf{fr}}}} 
        & 877 & 780 & 593 & 652 & 119 & 200 & 254 & 130 & 168 & 188 & 47 \\
        & 369 & 331 & 297 & 305 & 91 & 137 & 158 & 85 & 110 & 119 & 45 \\
        & 406 & 336 & 349 & 369 & 101 & 161 & 185 & 105 & 130 & 141 & 44 \\
    \hline 
    \parbox[t]{2mm}{\multirow{3}{*}{\rotatebox[origin=c]{90}{\textbf{sk}}}} 
        & 422 & 510 & 501 & 605 & 154 & 318 & 363 & 149 & 217 & 232 & 266 \\
        & 244 & 271 & 277 & 306 & 121 & 200 & 213 & 109 & 137 & 139 & 187 \\
        & 287 & 325 & 287 & 322 & 125 & 206 & 236 & 118 & 143 & 150 & 163 \\
    \hline
    \parbox[t]{2mm}{\multirow{3}{*}{\rotatebox[origin=c]{90}{\textbf{ta}}}} 
        & 129 & 133 & 112 & 112 & 61 & 65 & 65 & 33 & 33 & 33 & 117 \\
        & 94 & 94 & 79 & 79 & 54 & 61 & 61 & 29 & 29 & 29 & 63 \\
        & 119 & 117 & 78 & 78 & 48 & 57 & 58 & 26 & 26 & 26 & 77 \\
    \hline 
    \end{tabular}
    \caption{\label{tab:enc-stats}Number of labels ($|\mathcal{X}|)$ and dependency types ($|\mathcal{R}|$) produced with our encodings in each treebank.}
\end{table}

\subsection{Experimental setup}\label{ap:setup}
\begin{figure*}
    \centering
    \includegraphics[width=\textwidth]{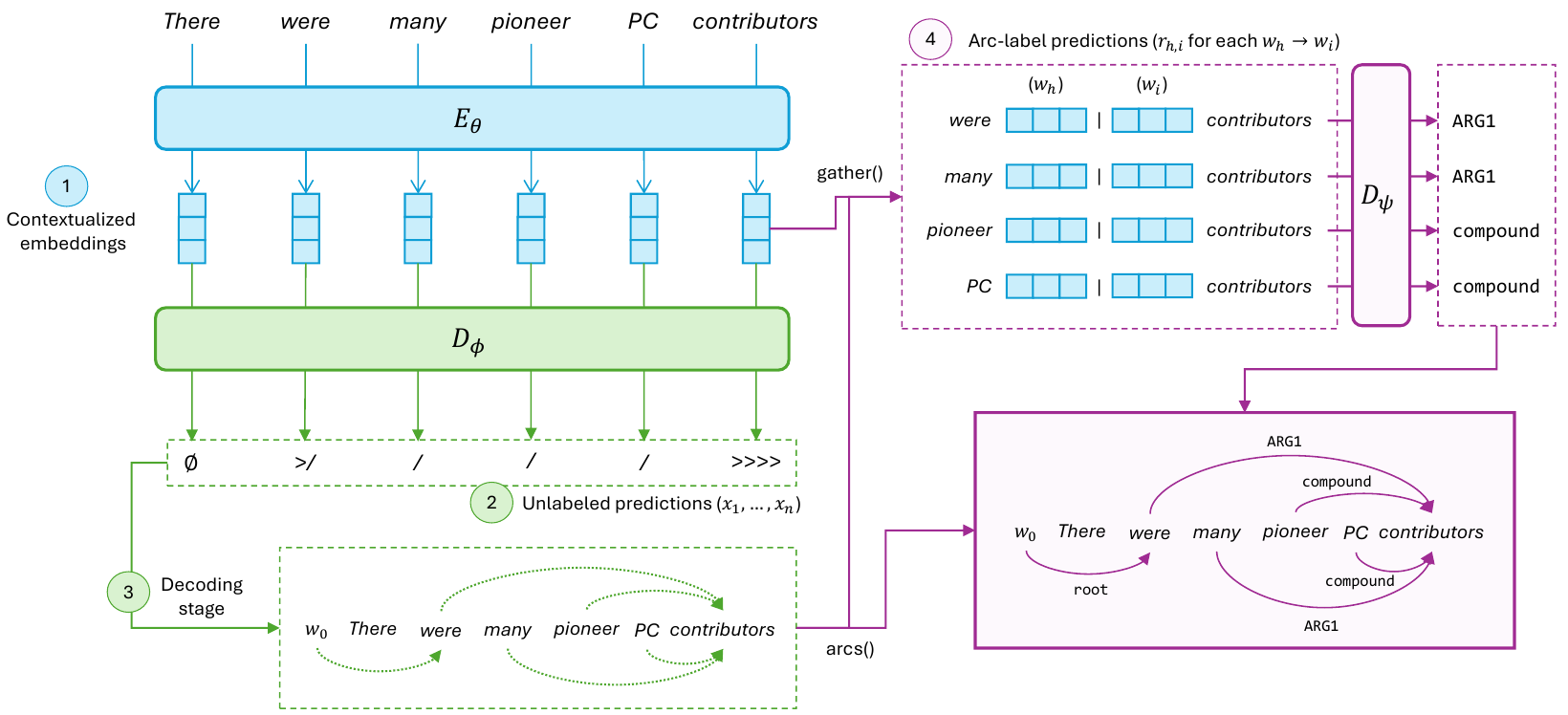}
    \caption{\label{fig:architecture}Prediction steps of our neural parser for the sentence ``\textit{There were many pioneer PC contributors}'' (from the DM\textsubscript{en} treebank) using the 1-planar bracketing encoding.}
\end{figure*}

Our linearization systems are conformed by two modules: a neural encoder $E_\theta:\mathbb{R}^{n\times d_x} \to\mathbb{R}^{n\times d_h} $ and a neural decoder with two different submodules: $D_\phi:\mathbb{R}^{d_h} \to L$ and $D_\varphi:\mathbb{R}^{2\cdot d_h}\to \mathcal{R}$.\footnote{Here we denote with $\mathcal{R}$ the set of dependency relationships between the edges of the graph.} Given a sentence $(w_1,...,w_n)$, the encoder $E_\theta$ outputs their contextualized representations ($\mathbf{w}_1, ...,\mathbf{w}_n$). Then, $D_\phi$ produces a label component $x_i\in \mathcal{X}$ for each hidden representation $\mathbf{w}_i$, and the decoding process recovers the set of predicted arcs $\hat{E}$ $\subseteq$ $\{ (w_h, w_i): h,i\in[1,n], h\neq i\}$. Finally, to obtain the type of dependency relation associated to each arc, $D_\varphi$ concatenates the word contextualizations of each predicted arc $(\mathbf{w}_h,\mathbf{w}_i)$ and outputs the dependency type $r_{h,i}\in \mathcal{R}$. Note that in this formulation, the component $d_i$ of the tuple $l_i=(d_i, x_i)$ introduced in Section \ref{sec:background} is not directly predicted with $D_\psi$, but the dependency relation associated to each decoded arc. One could define the component $d_i$ by sorting the elements of the set $\{r_{h,i}: (w_h, w_i)\in\hat{E}\}$ by $h$, and use $D_\psi$ to predict the entire sorted sequence. In practice, our implementation does not reconstruct $d_i$ as we just use each obtained $r_{h,i}$ to directly label the corresponding dependency. Figure \ref{fig:architecture} shows an illustration of the inference steps to reconstruct the labeled semantic graph for a given sentence. 

As a pretrained encoder, we primary finetuned XLM-RoBERTa-R \cite{conneau-etal-2020-unsupervised}, and XLNet \cite{yang2019xlnet}, for experiments in English. Additionally, we conducted experiments with a non-pretrained 2-layer BiLSTM encoder. We excluded these results from the main content of the paper, except for reporting accuracy-speed trade-offs, due to space constraints. However, we included them in Appendix \ref{ap:results} for a more complete picture.
The pretrained encoders were only fed with word information, excluding PoS-tags or character embeddings from the input, and a feed-forward network was stacked in the last layer of the architecture to reduce the original output dimension. For the decoder, we relied on a 1-layered feed-forward network that parameterizes each part  ($D_\phi$ and $D_\varphi$) of the module.

The full system was trained end-to-end with the AdamW optimizer \cite{loshchilov2019decoupled}, setting the learning rate $\eta$ to $1\cdot 10^{-3}$ or $1\cdot 10^{-4}$ for non-pretrained models and $5\cdot 10^{-5}$ or $1\cdot 10^{-5}$ for the pretrained architectures. All layers use a LeakyReLU \cite{xu2015empirical} with a negative slope of 0.01 as the activation function and the latent space applies a dropout of 0.33. Table \ref{tab:config} summarizes the hyperparameter selection to configure our neural models.

\begin{table}
    \centering\footnotesize
    \renewcommand{\arraystretch}{1.3}
    \begin{tabular}{c|c}
        \hline 
         \textbf{Hyp.} & \textbf{Model configuration} \\
         \hline 
         $d_h$ &  400 \\ 
         $d_w$ & 400 \\ 
         $d_p$ & 100 \\ 
         $d_c$ & 30 \\ 
        \hline 
         \textbf{Hyp.} & \textbf{Training configuration} \\
         \hline 
         epochs & 200 (BiLSTM), 200 (pretrained) \\ 
         batch size & 2000 (BiLSTM), 200 (pretrained) \\ 
         $\eta$ & \makecell[c]{$1\cdot 10^{-3}$ (BiLSTM),\\ $5\cdot 10^{-5}$ (XLM-R), $1\cdot 10^{-5}$ (XLNet)}\\ 
         \hline 
    \end{tabular}
    \caption{\label{tab:config}Model and training hyperparameters.}
\end{table}

\subsection{Results}\label{ap:results}
\begin{figure*}[h]
    \begin{subfigure}{0.48\textwidth}
        \caption{PAS (English) in-distribution set.}
        \includegraphics[width=\textwidth]{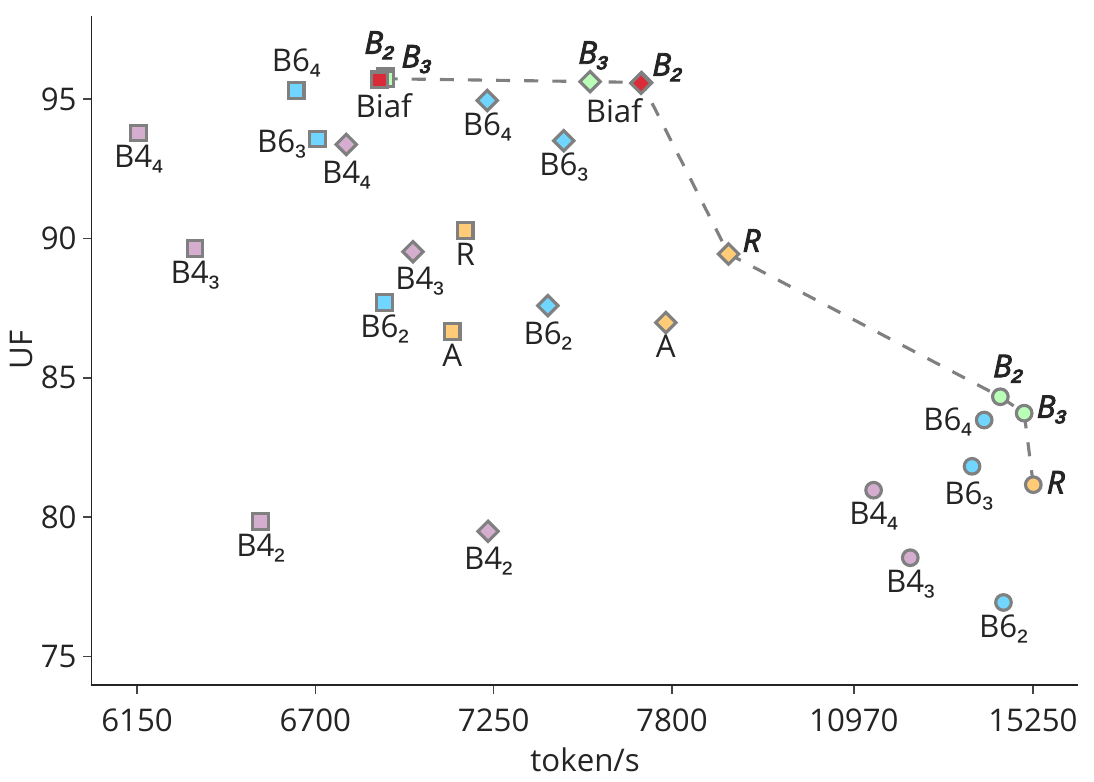}
    \end{subfigure}
    \begin{subfigure}{0.48\textwidth}
        \caption{PSD (English) in-distribution set.}
        \includegraphics[width=\textwidth]{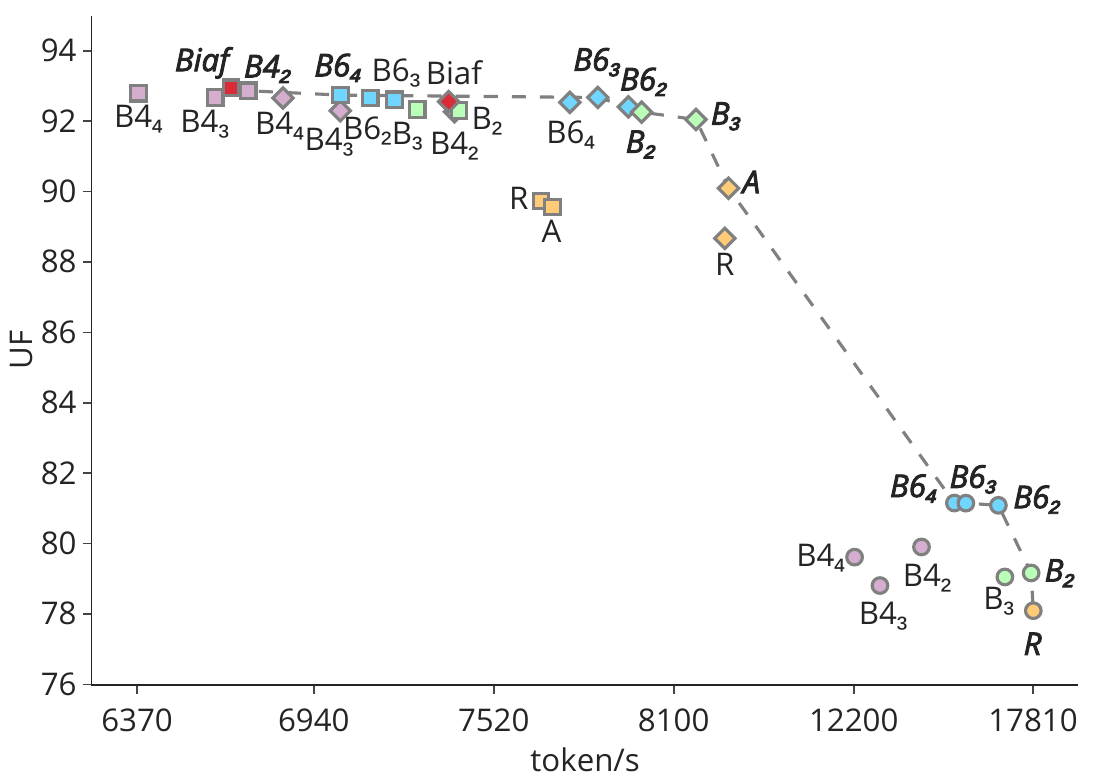}
    \end{subfigure}
    \begin{subfigure}{0.48\textwidth}
        \caption{PSD (Czech) test set.}
        \includegraphics[width=\textwidth]{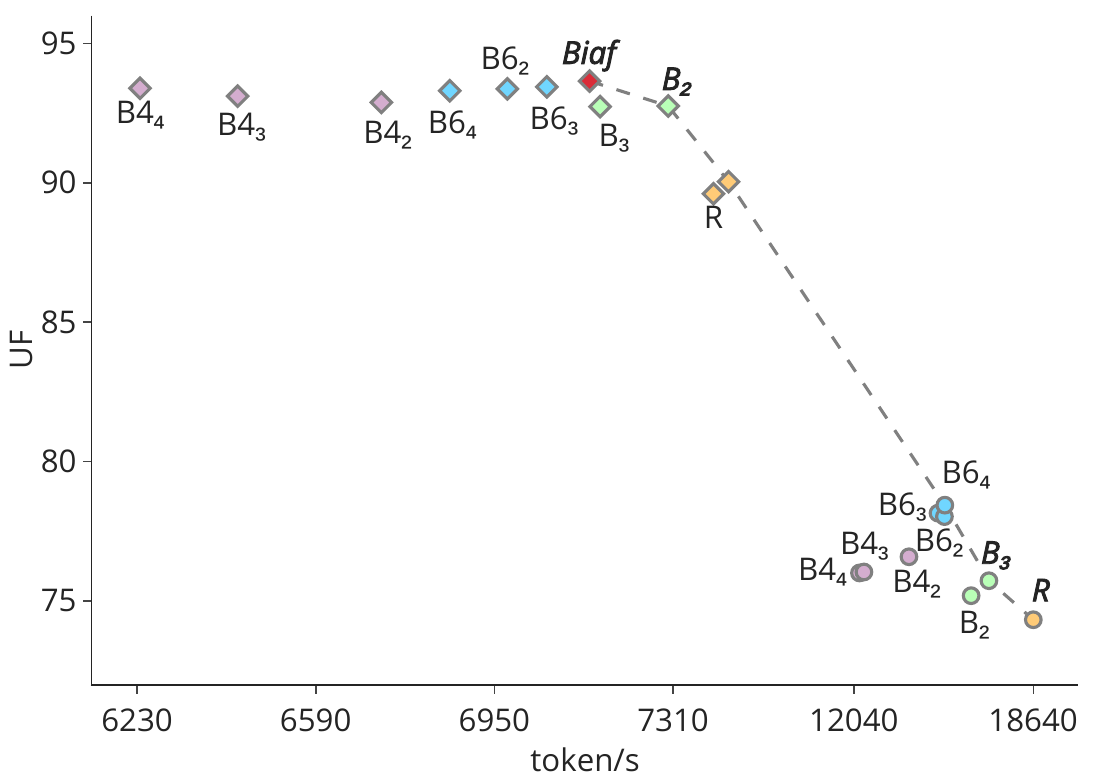}
    \end{subfigure}
    \begin{subfigure}{0.48\textwidth}
        \caption{PAS (Chinese) test set.}
        \includegraphics[width=\textwidth]{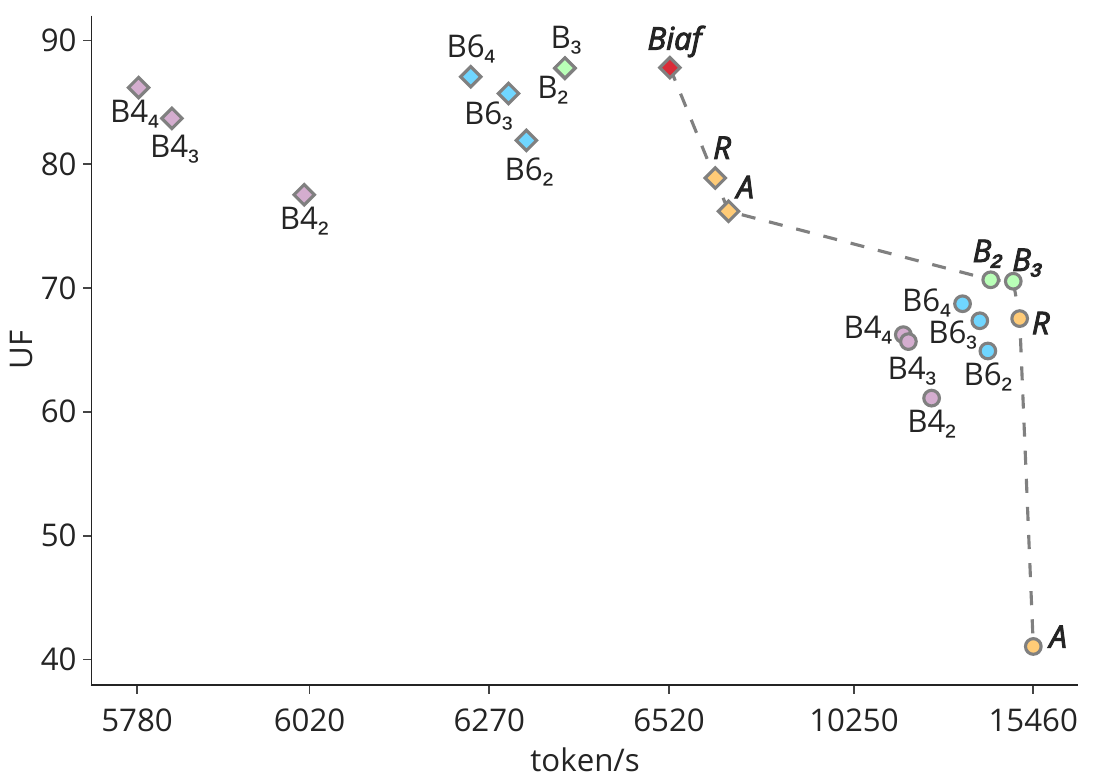}
    \end{subfigure}
    \begin{subfigure}{0.48\textwidth}
        \caption{Finnish-TDT/PUD test set.}
        \includegraphics[width=\textwidth]{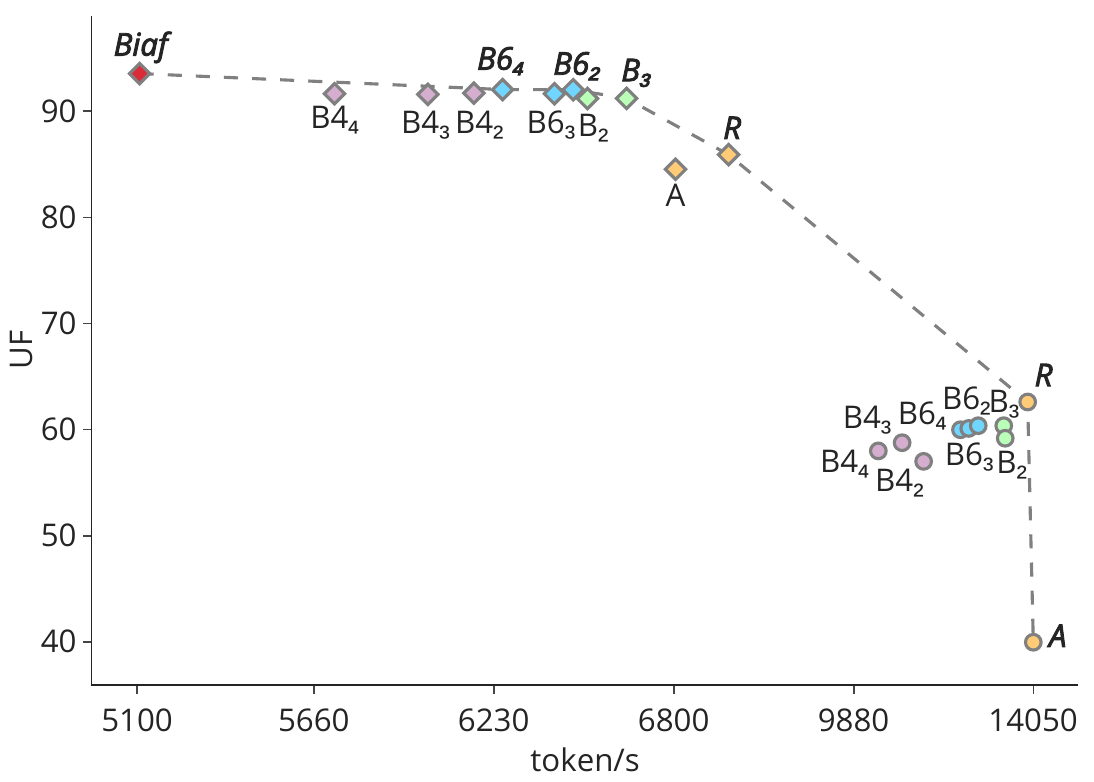}
    \end{subfigure}
    \begin{subfigure}{0.48\textwidth}
        \caption{Slovak-SNK test set.}
        \includegraphics[width=\textwidth]{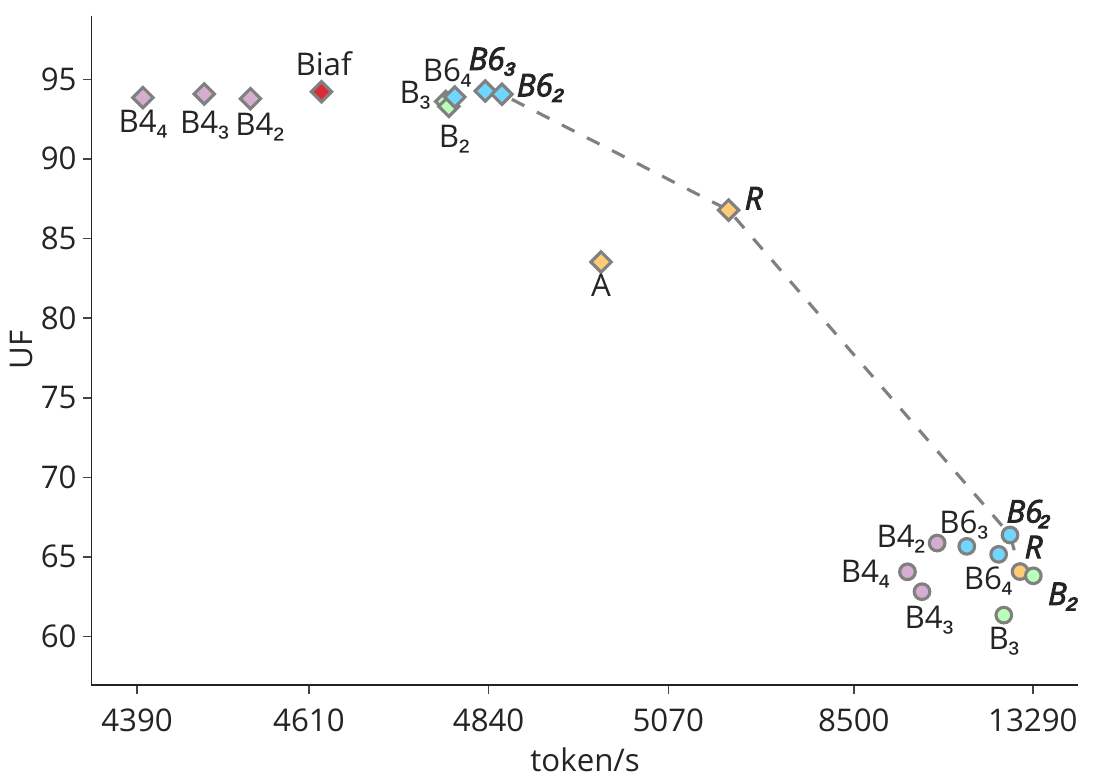}
    \end{subfigure}
    \begin{subfigure}{0.48\textwidth}
        \caption{Tamil-TTB test set.}
        \includegraphics[width=\textwidth]{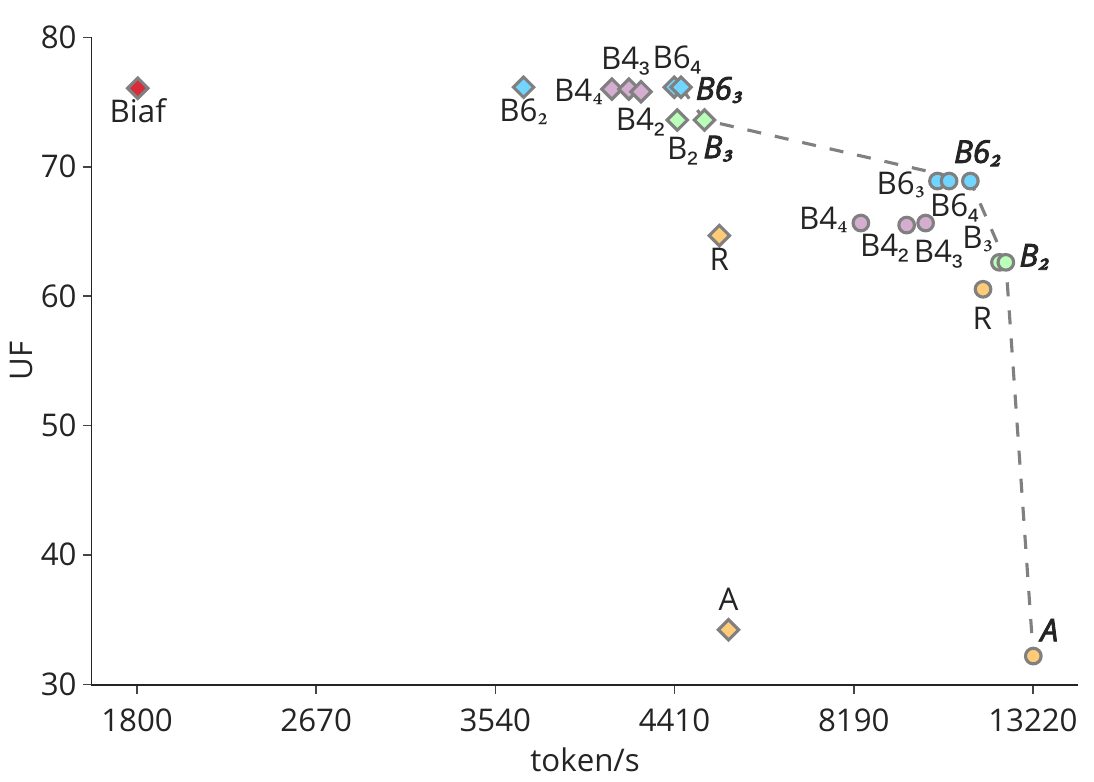}
    \end{subfigure}
    \caption{\label{fig:pareto2}Pareto front for the rest of treebanks.}
\end{figure*}

We report here more detailed results. Table~\ref{tab:dag-ood-results} presents the results for the out-of-domain DAG datasets, with trends similar to those observed for in-domain data in Table~\ref{tab:dag-id-results}. Tables~\ref{tab:dm-results} to \ref{tab:ta-results} show more detailed results for each dataset.
Metrics include both labeled and unlabeled F1 score (LF, UM), w.r.t. the predicted dependency (predicate, role, argument triplets); as well as labeled and unlabeled exact match, known as LM and UM. Additionally, we include the performance in terms of (i) tagging accuracy and ratio of well-formed trees in Table \ref{tab:acc}, (ii) number of planes in Table \ref{tab:planes} and (iii) number of cycles in Table \ref{tab:cycles}. Finally, Figure \ref{fig:pareto2} summarizes the rest of accuracy-speed comparisons for the treebanks not included in Section \ref{subsec:speed}.

\begin{table*}
    \centering\small
    \setlength{\tabcolsep}{3.2pt}
    \renewcommand{\arraystretch}{1.1}
\begin{tabular}{|c|cc|c|cc|c|cc|c|cc|c|}
    \hline 
                        & \multicolumn{3}{c|}{\bf DM\textsubscript{en}}                         &  \multicolumn{3}{c|}{\bf PAS\textsubscript{en}}                       & \multicolumn{3}{c|}{\bf PSD\textsubscript{en}}                        & \multicolumn{3}{c|}{\bf PSD\textsubscript{cs}}                        \\
    \cline{2-13} 
                        & \makecell[c]{\bf UF}  & \makecell[c]{\bf LF}  & \makecell[c]{\bf OF}  & \makecell[c]{\bf UF}  & \makecell[c]{\bf LF}  & \makecell[c]{\bf OF}  & \makecell[c]{\bf UF}  & \makecell[c]{\bf LF}  & \makecell[c]{\bf OF}  & \makecell[c]{\bf UF}  & \makecell[c]{\bf LF}  & \makecell[c]{\bf OF}  \\
    \hline 
    A                   & 87.47                 & 86.35                 & 100                   & 86.22                 & 84.62                 & 100                   & 88.55                 & 78.24                 & 100                   &86.92                  & 71.78                 & 100                    \\       
    R                   & 89.94                 & 88.90                 & 100                   & 90.14                 & 88.51                 & 100                   & 88.47                 & 78.26                 & 100                   &86.43                  & 71.43                 & 100                    \\                  
    B\textsubscript{2}  & 92.56                 & 91.54                 & 99.97                 & \best{94.82}          & \best{93.18}          & 99.96                 & 91.45                 & 81.38                 & 99.82                 &89.69                  & 74.89                 & 99.79                  \\
\hline 
\end{tabular}

    \caption{\small\label{tab:dag-ood-results}DAG out-of-distribution results. Notation as in Table~\ref{tab:dag-id-results}.}
\end{table*}

\begin{table*}
    \centering\scriptsize
    \setlength{\tabcolsep}{2pt}
    \renewcommand{\arraystretch}{1.1}
    \begin{tabular}{c|cccc|cccc|cccc}
    \hline
    & \multicolumn{4}{c|}{\textbf{dev}} & \multicolumn{4}{c|}{\textbf{in-distribution}} & \multicolumn{4}{c}{\textbf{out-of-distribution}} \\
    \hline
    \textbf{BiLSTM} & \textbf{UF} & \textbf{UM} & \textbf{LF} & \textbf{LM} & \textbf{UF} & \textbf{UM} & \textbf{LF} & \textbf{LM} & \textbf{UF} & \textbf{UM} & \textbf{LF} & \textbf{LM} \\
    \hline
    A & 59.38 & 4.61 & 57.07 & 4.33 & 57.85 & 6.98 & 54.45 & 6.38 & 60.15 & 5.32 & 57.99 & 4.73 \\
    R & 78.70 & 8.87 & 75.73 & 7.23 & 74.18 & 12.49 & 69.79 & 10.49 & 79.56 & 10.22 & 76.74 & 8.81 \\
    B\textsubscript{2}  & 80.77 & \textbf{13.76} & 78.36 & \textbf{12.34} & \textbf{75.83} & \textbf{15.14} & 71.93 & \textbf{12.82} & 81.81 & \textbf{15.25} & 79.25 & \textbf{13.65} \\
    B\textsubscript{3}  & \textbf{81.32} & 13.26 & \textbf{78.84} & 11.49 & 75.57 & 13.90 & 72.01 & 11.68 & \textbf{82.06} & 13.53 & \textbf{79.58} & 11.88 \\
    B4\textsubscript{2} & 73.11 & 4.47 & 71.03 & 4.04 & 69.71 & 8.98 & 67.07 & 7.79 & 73.27 & 4.43 & 71.15 & 4.08 \\
    B4\textsubscript{3} & 79.01 & 8.16 & 76.56 & 6.95 & 73.99 & 12.28 & 70.70 & 11.20 & 79.49 & 8.27 & 77.10 & 7.33 \\
\hline
\end{tabular}

    \caption{\label{tab:dm-results}Performance in the DM dev, in-distribution and out-of-distribution sets. OC is the oracle coverage and UF (LF) and UM (LM) are the unlabeled (labeled) F-score and exact match. The best performing model is highlighted in bold.}
\end{table*}

\begin{table*}
    \centering\scriptsize
    \setlength{\tabcolsep}{2pt}
    \renewcommand{\arraystretch}{1.1}
    \begin{tabular}{c|cccc|cccc|cccc}
    \hline
    & \multicolumn{4}{c|}{\textbf{dev}} & \multicolumn{4}{c|}{\textbf{in-distribution}} & \multicolumn{4}{c}{\textbf{out-of-distribution}} \\
    \hline
    \textbf{BiLSTM} & \textbf{UF} & \textbf{UM} & \textbf{LF} & \textbf{LM} & \textbf{UF} & \textbf{UM} & \textbf{LF} & \textbf{LM} & \textbf{UF} & \textbf{UM} & \textbf{LF} & \textbf{LM} \\
    \hline
    A & 65.52 & 4.40 & 63.88 & 4.33 & 63.07 & 6.65 & 60.58 & 6.33 & 65.89 & 5.61 & 64.00 & 5.32 \\
    R & 81.17 & 6.81 & 79.04 & 6.03 & 77.51 & 11.79 & 74.68 & 11.25 & 81.86 & 9.22 & 79.83 & 8.87 \\
    B\textsubscript{2} & \textbf{84.32} & \textbf{13.33} & \textbf{82.39} & \textbf{12.7} & \textbf{80.32} & \textbf{17.36} & \textbf{77.59} & \textbf{16.39} & \textbf{84.93} & \textbf{16.43} & \textbf{83.02} & \textbf{15.37} \\
    B\textsubscript{3} & 83.73 & 12.41 & 81.86 & 11.56 & 79.73 & 16.23 & 76.87 & 15.36 & 84.43 & 14.89 & 82.50 & 14.24 \\
    B4\textsubscript{2} & 69.78 & 1.21 & 68.08 & 1.13 & 68.85 & 4.87 & 66.50 & 4.76 & 70.18 & 2.31 & 68.54 & 2.31 \\
    B4\textsubscript{3} & 78.54 & 5.46 & 76.77 & 5.25 & 75.32 & 10.49 & 72.81 & 10.11 & 79.08 & 5.56 & 77.30 & 5.38 \\
\hline
\end{tabular}

    \caption{\label{tab:pas-results}Performance in the PAS (English) dataset. Notation as in Table \ref{tab:dm-results}.}
\end{table*}

\begin{table*}
    \centering\scriptsize
    \setlength{\tabcolsep}{3pt}
    \renewcommand{\arraystretch}{1.1}
    \begin{tabular}{c|cccc|cccc|cccc}
    \hline
    & \multicolumn{4}{c|}{\textbf{dev}} & \multicolumn{4}{c|}{\textbf{in-distribution}} & \multicolumn{4}{c}{\textbf{out-of-distribution}} \\
    \hline
    \textbf{BiLSTM} & \textbf{UF} & \textbf{UM} & \textbf{LF} & \textbf{LM} & \textbf{UF} & \textbf{UM} & \textbf{LF} & \textbf{LM} & \textbf{UF} & \textbf{UM} & \textbf{LF} & \textbf{LM} \\
    \hline
    A & 65.53 & 13.40 & 52.51 & 4.75 & 65.09 & 20.93 & 50.21 & 9.57 & 65.97 & 12.41 & 53.15 & 5.38 \\
    R & 78.09 & 15.11 & 62.44 & 5.60 & 74.78 & 23.69 & 58.79 & 10.55 & 79.01 & 16.84 & 63.98 & 7.09 \\
    B\textsubscript{2} & 79.17 & 16.52 & 64.83 & 5.96 & 76.42 & 24.99 & 61.99 & 11.47 & 80.30 & 17.26 & 66.45 & 7.74 \\
    B\textsubscript{3} & 79.05 & 15.39 & 64.24 & 5.39 & 76.25 & 23.15 & 61.13 & 11.03 & 79.88 & 16.19 & 65.31 & 6.44 \\
    B4\textsubscript{2} & 79.90 & 17.09 & 66.36 & 6.10 & 76.19 & 25.74 & 62.75 & 12.71 & 80.64 & 17.67 & 68.25 & 7.56 \\
    B4\textsubscript{3} & 78.81 & 17.16 & 65.00 & 5.82 & 76.04 & 25.15 & 62.44 & 12.44 & 80.01 & 17.26 & 67.09 & 7.51 \\
\hline
\end{tabular}

    \caption{\label{tab:psd-results}Performance in the PSD (English) dataset. Notation as in Table \ref{tab:dm-results}.}
\end{table*}
\begin{table*}
    \centering\scriptsize
    \setlength{\tabcolsep}{3pt}
    \renewcommand{\arraystretch}{1.1}
    \begin{tabular}{c|cccc|cccc|cccc}
    \hline
    & \multicolumn{4}{c|}{\textbf{dev}} & \multicolumn{4}{c|}{\textbf{in-distribution}} & \multicolumn{4}{c}{\textbf{out-of-distribution}} \\
    \hline
    \textbf{BiLSTM} & \textbf{UF} & \textbf{UM} & \textbf{LF} & \textbf{LM} & \textbf{UF} & \textbf{UM} & \textbf{LF} & \textbf{LM} & \textbf{UF} & \textbf{UM} & \textbf{LF} & \textbf{LM} \\
    \hline
    A & 56.59 & 8.02 & 48.84 & 4.55 & 53.72 & 12.78 & 38.43 & 7.02 & 58.19 & 8.21 & 50.62 & 4.73 \\
    R & 74.32 & 11.86 & 63.72 & 6.11 & 68.16 & 15.98 & 48.51 & \textbf{8.02} & 76.36 & 11.69 & 66.70 & 6.57 \\
    B\textsubscript{2} & 75.18 & 10.30 & 65.30 & 5.45 & 67.94 & 14.33 & 48.78 & 7.16 & 76.95 & 11.79 & 67.73 & 6.37 \\
    B\textsubscript{3} & 75.72 & 10.96 & 66.35 & 5.81 & 68.74 & 14.87 & 50.21 & 7.54 & 77.68 & 11.64 & 68.67 & 6.22 \\
    B4\textsubscript{2} & 76.58 & 11.92 & 66.89 & 6.05 & 68.48 & 16.34 & 50.68 & 7.96 & 77.79 & 13.38 & 69.23 & 7.26 \\
    B4\textsubscript{3} & 76.03 & 12.99 & 66.50 & 5.81 & 67.95 & 15.63 & 49.73 & 7.67 & 77.37 & 13.23 & 68.75 & 7.06 \\
\hline
\end{tabular}

    \caption{\label{tab:psd-cs-results}Performance in the PSD (Czech) dataset. Notation as in Table \ref{tab:dm-results}.}
\end{table*}
\begin{table}
    \centering\scriptsize
    \setlength{\tabcolsep}{3pt}
    \renewcommand{\arraystretch}{1.1}
    \begin{tabular}{c|cccc|cccc}
    \hline
    & \multicolumn{4}{c|}{\textbf{dev}} & \multicolumn{4}{c}{\textbf{in-distribution}} \\
    \hline
    \textbf{BiLSTM} & \textbf{UF} & \textbf{UM} & \textbf{LF} & \textbf{LM} & \textbf{UF} & \textbf{UM} & \textbf{LF} & \textbf{LM} \\
    \hline
    A & 41.06 & 5.72 & 39.56 & 5.66 & 39.31 & 8.40 & 37.80 & 8.20 \\
    R & 67.54 & 7.79 & 65.01 & 7.49 & 67.13 & 9.71 & 64.47 & 9.51 \\
    B\textsubscript{2} & \textbf{70.67} & 11.41 & \textbf{68.58} & 11.27 & 68.87 & \textbf{11.8} & 66.54 & \textbf{11.72} \\
    B\textsubscript{3} & 70.55 & \textbf{11.81} & 68.41 & \textbf{11.7} & \textbf{69.06} & 11.76 & \textbf{66.72} & 11.68 \\
    B4\textsubscript{2} & 61.11 & 3.64 & 59.21 & 3.62 & 59.52 & 2.50 & 57.51 & 2.50 \\
    B4\textsubscript{3} & 65.68 & 7.64 & 63.64 & 7.59 & 64.29 & 7.17 & 62.09 & 7.05 \\
\hline
\end{tabular}

    \caption{\label{tab:pas-zh-results}Performance in the PAS (Chinese) dataset. Notation as in Table \ref{tab:dm-results}.}
\end{table}

\begin{table}
    \centering\scriptsize
    \setlength{\tabcolsep}{3pt}
    \renewcommand{\arraystretch}{1.1}
    \begin{tabular}{c|cccc|cccc}
    \hline
    & \multicolumn{4}{c|}{\textbf{dev}} & \multicolumn{4}{c}{\textbf{test}} \\
    \hline
    \textbf{BiLSTM} & \textbf{UF} & \textbf{UM} & \textbf{LF} & \textbf{LM} & \textbf{UF} & \textbf{UM} & \textbf{LF} & \textbf{LM} \\
    \hline
    A & 46.97 & 5.59 & 42.18 & 4.56 & 51.70 & 1.87 & 46.45 & 1.21 \\
    R & 71.88 & 6.03 & 63.73 & 4.85 & 71.30 & 2.53 & 63.46 & 1.32 \\
    B\textsubscript{2} & 74.66 & 6.03 & 67.32 & 4.56 & 73.86 & 3.08 & 66.57 & 1.98 \\
    B\textsubscript{3} & 75.49 & \textbf{7.21} & 68.46 & 5.44 & 74.27 & 3.41 & 67.26 & 1.76 \\
    B4\textsubscript{2} & 74.52 & 6.18 & 66.82 & 4.71 & 73.25 & 3.41 & 65.80 & \textbf{2.31} \\
    B4\textsubscript{3} & 75.77 & 6.18 & 68.57 & 5.00 & 74.37 & 3.41 & 67.12 & 1.98 \\
\hline
\end{tabular}

    \caption{\label{tab:ar-results}Performance in the IWPT (Arabic) dataset. Notation as in Table \ref{tab:dm-results}.}
\end{table}
\begin{table}
    \centering\scriptsize
    \setlength{\tabcolsep}{3pt}
    \renewcommand{\arraystretch}{1.1}
    \begin{tabular}{c|cccc|cccc}
    \hline
    & \multicolumn{4}{c|}{\textbf{dev}} & \multicolumn{4}{c}{\textbf{test}} \\
    \hline
    \textbf{BiLSTM} & \textbf{UF} & \textbf{UM} & \textbf{LF} & \textbf{LM} & \textbf{UF} & \textbf{UM} & \textbf{LF} & \textbf{LM} \\
    \hline
    A & 39.97 & 5.83 & 33.29 & 3.33 & 41.89 & 10.41 & 34.84 & 7.48 \\
    R & \textbf{62.6} & \textbf{9.43} & \textbf{49.33} & 3.56 & \textbf{62.93} & \textbf{14.81} & 49.87 & \textbf{8.06} \\
    B\textsubscript{2} & 59.19 & 7.36 & 48.40 & \textbf{4.11} & 58.82 & 11.14 & 47.96 & 7.77 \\
    B\textsubscript{3} & 60.37 & 8.10 & 48.60 & 3.95 & 61.06 & 12.02 & 49.41 & 7.70 \\
    B4\textsubscript{2} & 57.01 & 6.38 & 46.09 & 3.13 & 57.68 & 9.46 & 46.44 & 6.01 \\
    B4\textsubscript{3} & 58.76 & 7.44 & 47.79 & 3.84 & 58.77 & 10.56 & 47.71 & 6.82 \\
\hline
\end{tabular}

    \caption{\label{tab:fi-results}Performance in the IWPT (Finnish) dataset.}
\end{table}
\begin{table}
    \centering\scriptsize
    \setlength{\tabcolsep}{3pt}
    \renewcommand{\arraystretch}{1.1}
    \begin{tabular}{c|cccc|cccc}
    \hline
    & \multicolumn{4}{c|}{\textbf{dev}} & \multicolumn{4}{c}{\textbf{test}} \\
    \hline
    \textbf{BiLSTM} & \textbf{UF} & \textbf{UM} & \textbf{LF} & \textbf{LM} & \textbf{UF} & \textbf{UM} & \textbf{LF} & \textbf{LM} \\
    \hline
    A & 55.41 & 11.91 & 48.92 & 5.14 & 43.33 & 8.74 & 40.80 & 7.04 \\
    R & 73.91 & 8.82 & 64.81 & 6.01 & 75.00 & 10.92 & 69.98 & 8.50 \\
    B\textsubscript{2} & 76.62 & 19.09 & 67.18 & 4.95 & 78.92 & 11.41 & 74.04 & 9.95 \\
    B\textsubscript{3} & 76.79 & 17.70 & 67.96 & 5.79 & 78.94 & 11.16 & 74.05 & 10.19 \\
    B4\textsubscript{2} & 76.28 & 11.33 & 66.47 & 4.85 & 78.74 & 12.14 & 72.59 & 8.98 \\
    B4\textsubscript{3} & 76.21 & 9.62 & 66.67 & 4.70 & 78.92 & 10.44 & 72.88 & 7.04 \\
\hline
\end{tabular}

    \caption{\label{tab:fr-results}Performance in the IWPT (French) dataset. Notation as in Table \ref{tab:dm-results}.}
\end{table}
\begin{table}
    \centering\scriptsize
    \setlength{\tabcolsep}{3pt}
    \renewcommand{\arraystretch}{1.1}
    \begin{tabular}{c|cccc|cccc}
    \hline
    & \multicolumn{4}{c|}{\textbf{dev}} & \multicolumn{4}{c}{\textbf{test}} \\
    \hline
    \textbf{BiLSTM} & \textbf{UF} & \textbf{UM} & \textbf{LF} & \textbf{LM} & \textbf{UF} & \textbf{UM} & \textbf{LF} & \textbf{LM} \\
    \hline
    A & 40.75 & 12.54 & 33.47 & 8.67 & 41.63 & 12.17 & 33.93 & 7.83 \\
    R & 64.08 & 17.06 & 51.48 & 11.59 & 65.08 & 17.55 & 52.64 & 10.09 \\
    B\textsubscript{2} & 63.82 & 17.15 & 51.66 & 9.80 & 64.62 & 17.45 & 52.96 & 10.19 \\
    B\textsubscript{3} & 61.35 & 17.15 & 50.15 & 11.78 & 61.97 & 16.98 & 50.64 & 10.19 \\
    B4\textsubscript{2} & 65.88 & 16.31 & \textbf{53.17} & 9.52 & 65.55 & 15.66 & 53.65 & 8.49 \\
    B4\textsubscript{3} & 62.82 & 14.89 & 51.34 & 11.12 & 63.37 & 14.62 & 51.94 & 8.49 \\
\hline
\end{tabular}

    \caption{\label{tab:sk-results}Performance in the IWPT (Slovak) dataset. Notation as in Table \ref{tab:dm-results}.}
\end{table}
\begin{table}
    \centering\scriptsize
    \setlength{\tabcolsep}{3pt}
    \renewcommand{\arraystretch}{1.1}
    \begin{tabular}{c|cccc|cccc}
    \hline
    & \multicolumn{4}{c|}{\textbf{dev}} & \multicolumn{4}{c}{\textbf{test}} \\
    \hline
    \textbf{BiLSTM} & \textbf{UF} & \textbf{UM} & \textbf{LF} & \textbf{LM} & \textbf{UF} & \textbf{UM} & \textbf{LF} & \textbf{LM} \\
    \hline
    A & 32.19 & 0.00 & 23.37 & 0.00 & 32.79 & 0.00 & 23.54 & 0.00 \\
    R & 60.54 & 0.83 & 41.71 & 0.00 & 60.99 & 0.00 & 44.96 & 0.00 \\
    B\textsubscript{2} & 62.62 & 4.17 & 42.91 & 0.00 & 65.52 & \textbf{6.25} & 47.71 & 2.50 \\
    B\textsubscript{3} & 62.62 & 4.17 & 42.91 & 0.00 & 65.52 & 6.25 & 47.71 & 2.50 \\
    B4\textsubscript{2} & 65.50 & 0.00 & 47.10 & 0.00 & 64.25 & 6.25 & 49.39 & 3.75 \\
    B4\textsubscript{3} & 65.66 & 3.33 & 48.34 & 0.00 & 62.96 & 5.00 & \textbf{49.86} & \textbf{5.0} \\
\hline
\end{tabular}

    \caption{\label{tab:ta-results}Performance in the IWPT (Tamil) dataset.}
\end{table}
\begin{table*}
    \centering\scriptsize
    \setlength{\tabcolsep}{2.5pt}
    \renewcommand{\arraystretch}{1.1}
    \begin{tabular}{|c|c|cccccccccc|cccccccccc|}
    \cline{3-22}
    \multicolumn{2}{c|}{}& \multicolumn{10}{c|}{\textbf{tag-accuracy}} & \multicolumn{10}{c|}{\textbf{\%well-formed}}\\
    \cline{3-22}
    \multicolumn{2}{c|}{}& \textbf{A} & \textbf{R} & \textbf{B\textsubscript{2}} & \textbf{B\textsubscript{3}} & \textbf{B4\textsubscript{2}} &\textbf{ B4\textsubscript{3}} & \textbf{B4\textsubscript{4}} & \textbf{B6\textsubscript{2}} & \textbf{B6\textsubscript{3}} & \textbf{B6\textsubscript{4}} & \textbf{A} & \textbf{R} & \textbf{B\textsubscript{2}} & \textbf{B\textsubscript{3}} & \textbf{B4\textsubscript{2}} &\textbf{ B4\textsubscript{3}} & \textbf{B4\textsubscript{4}} & \textbf{B6\textsubscript{2}} & \textbf{B6\textsubscript{3}} & \textbf{B6\textsubscript{4}}\\
    \cline{3-22}
    \multicolumn{2}{c|}{} & \multicolumn{20}{c|}{\textbf{DAG}} \\
    \hline 
    \parbox[t]{2mm}{\multirow{9}{*}{\rotatebox[origin=c]{90}{\textbf{DM\textsubscript{en}}}}}
    & \parbox[t]{2mm}{\multirow{3}{*}{\rotatebox[origin=c]{90}{\textbf{dev}}}} 
        & 72.31 & 81.61 & 78.73 & 78.77 & 78.03 & 73.18 & 71.33 & 80.45 & 79.60 & 79.22 & 24.82 & 92.20 & 15.31 & 14.81 & 47.12 & 33.99 & 29.22 & 38.59 & 36.60 & 38.22 \\
        && 92.02 & 92.45 & 88.36 & 88.45 & 93.84 & 92.28 & 91.40 & 93.97 & 93.63 & 93.64 & 79.61 & 86.46 & 78.67 & 81.19 & 94.73 & 87.50 & 88.76 & 87.71 & 81.74 & 87.30 \\
        && 92.11 & 93.02 & 88.54 & 88.34 & 94.52 & 92.20 & 91.99 & 94.55 & 94.15 & 94.19 & 76.07 & 89.66 & 78.89 & 70.03 & 91.61 & 81.30 & 91.63 & 91.70 & 89.65 & 89.13 \\
    \cline{2-22}
    & \parbox[t]{2mm}{\multirow{3}{*}{\rotatebox[origin=c]{90}{\textbf{id}}}}
        & 71.97 & 81.07 & 77.61 & 77.84 & 77.22 & 72.29 & 70.32 & 79.59 & 78.91 & 78.78 & 22.70 & 91.88 & 14.92 & 14.67 & 47.97 & 34.49 & 30.69 & 36.79 & 36.38 & 41.47 \\
        && 92.08 & 92.18 & 87.29 & 87.20 & 93.62 & 91.81 & 90.96 & 93.88 & 93.17 & 93.26 & 80.21 & 86.84 & 77.31 & 79.49 & 92.60 & 87.34 & 86.44 & 87.64 & 82.57 & 87.89 \\
\hline 
\end{tabular}
    \caption{\label{tab:acc}Performance in terms of tagging accuracy (\textbf{tag-accuracy}) and percentage of well-formed trees without post-processing (\textbf{\%well-formed}) for the DAG and IWPT treebanks. Each subrow determines the encoder used (BiLSTM, XLM and, for treebanks in English, XLNet). The average performance is also included ($\mu$).}
\end{table*}

\begin{table*}
    \centering\scriptsize
    \setlength{\tabcolsep}{2.1pt}
    \renewcommand{\arraystretch}{1.1}
    \begin{tabular}{|c|c|c|cccccccccc|cccccccccc|c|}
    \cline{4-24}
    \multicolumn{3}{c|}{}& \multicolumn{10}{c|}{\textbf{BiLSTM}} & \multicolumn{11}{c|}{\textbf{XLM | XLNet}}\\
    \cline{3-24}
    \multicolumn{2}{c|}{} &$k$& \textbf{A} & \textbf{R} & \textbf{B\textsubscript{2}} & \textbf{B\textsubscript{3}} & \textbf{B4\textsubscript{2}} &\textbf{ B4\textsubscript{3}} & \textbf{B4\textsubscript{4}} & \textbf{B6\textsubscript{2}} & \textbf{B6\textsubscript{3}} & \textbf{B6\textsubscript{4}} & \textbf{A} & \textbf{R} & \textbf{B\textsubscript{2}} & \textbf{B\textsubscript{3}} & \textbf{B4\textsubscript{2}} &\textbf{ B4\textsubscript{3}} & \textbf{B4\textsubscript{4}} & \textbf{B6\textsubscript{2}} & \textbf{B6\textsubscript{3}} & \textbf{B6\textsubscript{4}} & \textbf{Biaf}\\
    \hline
    \multicolumn{24}{|c|}{\textbf{DAG}} \\
    \hline 
    \parbox[t]{2mm}{\multirow{7}{*}{\rotatebox[origin=c]{90}{\textbf{DM\textsubscript{en}}}}}
    & \parbox[t]{2mm}{\multirow{4}{*}{\rotatebox[origin=c]{90}{\textbf{id}}}} 
        & 1 & 61.63 & 79.89 & 82.72 & 83.29 & 74.07 & 80.54 & 82.18 & 78.00 & 81.69 & 82.14 & 89.73 & 92.52 & 95.44 & 95.14 & 85.95 & 93.02 & 95.36 & 91.45 & 95.34 & 95.66 & 95.32 \\
        && 2 & 57.40 & 77.91 & 79.19 & 79.74 & 72.49 & 77.79 & 79.54 & 75.81 & 78.66 & 79.57 & 87.92 & 91.59 & 95.28 & 94.51 & 87.09 & 92.48 & 95.02 & 91.64 & 94.75 & 95.03 & 95.01 \\
        && 3 & 53.44 & 70.83 & 72.32 & 72.37 & 66.47 & 72.76 & 72.71 & 67.31 & 70.02 & 73.09 & 81.97 & 87.38 & 87.91 & 87.17 & 84.31 & 88.67 & 90.73 & 88.05 & 89.47 & 91.21 & 92.21 \\
        && 4 & 25.81 & 44.44 & 56.25 & 52.94 & 48.48 & 38.71 & 55.17 & 48.48 & 45.71 & 43.75 & 72.73 & 57.14 & 57.14 & 54.55 & 58.82 & 64.71 & 64.71 & 58.82 & 62.86 & 62.86 & 64.71 \\
\hline
\end{tabular}
    \caption{\label{tab:planes}UF score per plane in the DAG and IWPT evaluation treebanks. Each subrow indicates the score achieved for graphs with $k$ planes and the average performance across treebanks is included ($\mu$).}
\end{table*}

{\clearpage
\onecolumn
\centering\scriptsize
\setlength{\tabcolsep}{2.3pt}
\renewcommand{\arraystretch}{1.1}
\begin{longtable}{|c|c|cccccccccc|cccccccccc|c|}
    \cline{3-23}
    \multicolumn{2}{c|}{}& \multicolumn{10}{c|}{\textbf{BiLSTM}} & \multicolumn{11}{c|}{\textbf{XLM}}\\
    \cline{2-23}
    \multicolumn{1}{c|}{} &$m$& \textbf{A} & \textbf{R} & \textbf{B\textsubscript{2}} & \textbf{B\textsubscript{3}} & \textbf{B4\textsubscript{2}} &\textbf{ B4\textsubscript{3}} & \textbf{B4\textsubscript{4}} & \textbf{B6\textsubscript{2}} & \textbf{B6\textsubscript{3}} & \textbf{B6\textsubscript{4}} & \textbf{A} & \textbf{R} & \textbf{B\textsubscript{2}} & \textbf{B\textsubscript{3}} & \textbf{B4\textsubscript{2}} &\textbf{ B4\textsubscript{3}} & \textbf{B4\textsubscript{4}} & \textbf{B6\textsubscript{2}} & \textbf{B6\textsubscript{3}} & \textbf{B6\textsubscript{4}} & \textbf{Biaf}\\
    \hline 
    \parbox[t]{2mm}{\multirow{5}{*}{\rotatebox[origin=c]{90}{\textbf{ar}}}}
        & 0 & 51.23 & 72.10 & 74.56 & 75.53 & 74.51 & 75.84 & 75.53 & 67.66 & 75.95 & 76.21 & 78.21 & 82.32 & 87.67 & 87.63 & 87.65 & 87.67 & 87.89 & 87.74 & 88.11 & 88.27 & 89.53 \\
        & 1 & 43.48 & 71.99 & 75.41 & 76.10 & 74.98 & 76.19 & 75.63 & 67.53 & 75.59 & 76.63 & 75.26 & 82.08 & 88.54 & 88.48 & 88.55 & 88.59 & 88.69 & 88.11 & 88.40 & 88.94 & 90.68 \\
        & 2 & 38.03 & 71.38 & 74.54 & 75.27 & 73.88 & 74.87 & 74.58 & 68.44 & 75.00 & 75.23 & 62.28 & 81.59 & 88.06 & 87.73 & 88.16 & 87.09 & 87.67 & 86.69 & 87.72 & 87.94 & 88.77 \\
        & 3 & 30.52 & 69.90 & 71.93 & 71.09 & 72.43 & 72.42 & 72.13 & 64.79 & 72.77 & 74.39 & 56.79 & 81.28 & 85.88 & 87.04 & 86.06 & 85.79 & 87.99 & 85.92 & 86.75 & 86.51 & 88.82 \\
        & 4 & 25.86 & 68.63 & 71.30 & 72.94 & 73.56 & 76.00 & 73.85 & 60.65 & 74.23 & 73.58 & 47.50 & 76.41 & 85.16 & 85.14 & 84.63 & 86.42 & 86.70 & 85.90 & 86.81 & 85.11 & 86.20 \\
\end{longtable}
}

\end{document}